%% file: main.tex
\documentclass[10pt,twocolumn,letterpaper]{article}

\usepackage{cvpr}              % To produce the CAMERA-READY version
\input{preamble}

\definecolor{cvprblue}{rgb}{0.21,0.49,0.74}
\usepackage[pagebackref,breaklinks,colorlinks,allcolors=cvprblue]{hyperref}

\def\paperID{1809} % *** Enter the Paper ID here
\def\confName{CVPR}
\def\confYear{2025}

\title{Seeing the Abstract: \\Translating the Abstract Language for Vision Language Models}

\author{Davide Talon*\textsuperscript{$1$} \quad
Federico Girella*\textsuperscript{$2$} \quad Ziyue Liu\textsuperscript{$2, 3$} \quad Marco Cristani\textsuperscript{$2$} 
\vspace{5pt}
\quad Yiming Wang\textsuperscript{$1$}\\
\textsuperscript{$1$}Fondazione Bruno Kessler\\
\textsuperscript{$2$}University of Verona \\
\vspace{5pt}
\textsuperscript{$3$}Polytechnic Institute of Turin \\
\url{https://github.com/davidetalon/fashionact}
}

\begin{document}

\twocolumn[{%
\renewcommand\twocolumn[1][]{#1}%
\maketitle
}]

\maketitle
\input{sec/0_abstract} 

\begin{figure}[t!]
    \centering
    \includegraphics[width=0.94\linewidth]{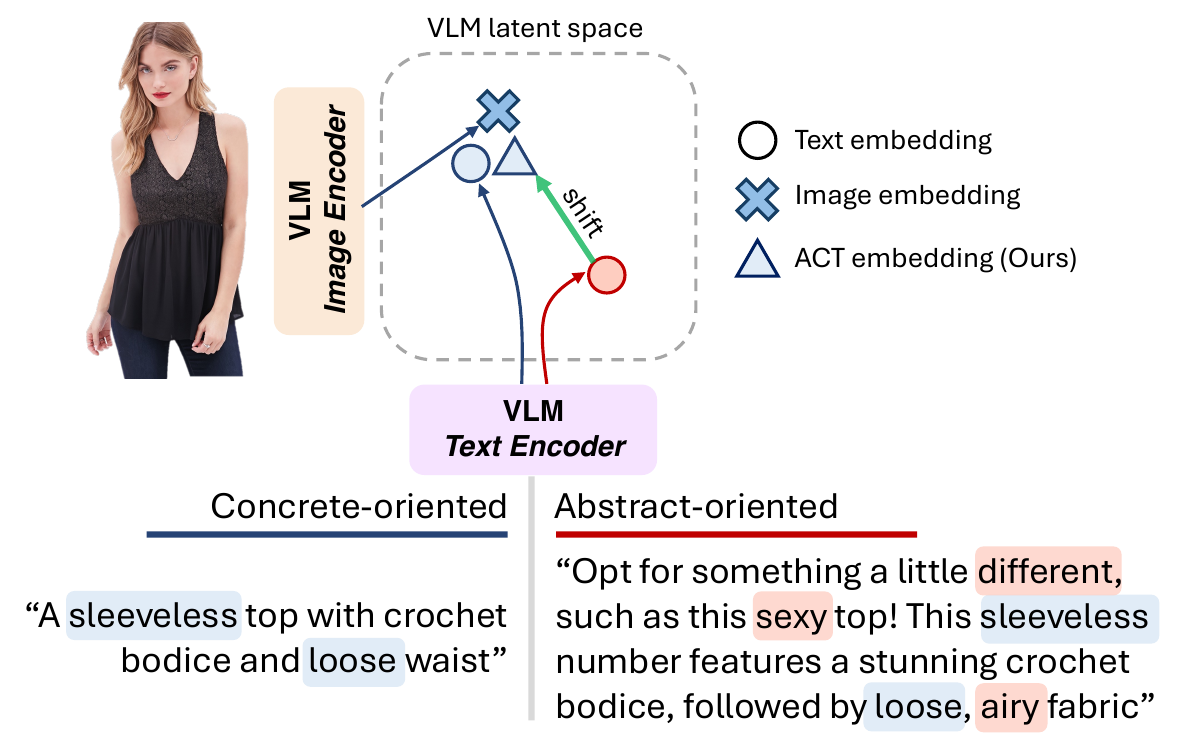}
    \caption{Human language can exhibit both \inlineColorbox{AbstractRed}{\emph{abstract}} and \inlineColorbox{ConcreteBlue}{\emph{concrete}} words to express feelings, desires, and properties together with perceivable elements, \eg when describing a fashion item. However, Vision Language Models (VLMs) are mostly pre-trained with \inlineColorbox{ConcreteBlue}{\emph{concrete}}-oriented web-image texts, thus under-representing the abstract-oriented ones.
    When encoding the abstract-oriented description with pre-trained VLMs, there exists a noticeable representation shift from the concrete-oriented description, hindering the performance in downstream tasks, \eg text-to-image retrieval in fashion.
    Our proposal \methodfull (\methodshort) can effectively compensate for such representation shift in a training-free manner, bringing the representation of the abstract-oriented language towards the concrete-oriented one in the latent space of existing VLMs, thereby improving the downstream task performance.}
    
    \label{fig:teaser}
\end{figure}

\footnotetext[1]{Equal Contribution}
\input{sec/1_intro}
\input{sec/2_related}
\input{sec/3_preliminary}
\input{sec/4_method}
\input{sec/5_experiments}
\input{sec/6_conclusions}
\input{sec/acknowledgements}
%\clearpage
{
    \small
    \bibliographystyle{ieeenat_fullname}
    \bibliography{main}
}
\input{supp/supp-preamble}

\input{supp/intro}
\input{supp/further-details}
\input{supp/additional-exp}
\input{supp/qualitatives}

%   \appendix
%\input{supp/plug-n-play}   


\end{document}

%% file: preamble.tex
\usepackage[normalem]{ulem}
\usepackage{booktabs}
\usepackage{multirow}
\usepackage{subcaption}
\usepackage{svg}
\usepackage{makecell}

\usepackage{pifont}
\usepackage{arydshln}
\usepackage{multirow}
\usepackage{makecell}
\usepackage{alphalph}
\usepackage{wrapfig}
\definecolor{tab-green}{HTML}{dcedc1}
\definecolor{tab-red}{HTML}{ffd3b6}
\definecolor{tab-blue}{HTML}{bae1ff}
\definecolor{refs-blue}{HTML}{367dbd}

\newcommand{\boldedparagraph}[1]{\smallskip

\noindent\textbf{#1}}

\def\ours{\methodshort}
\def\methodfull{Abstract-to-Concrete Translator\xspace}
\def\methodshort{ACT\xspace}

\def\AC{A-C\xspace}
\def\eg{\textit{e.g.,}\xspace} 
\def\ie{\textit{i.e.,}\xspace}

\def\suppmat{\textit{Supp. Mat.}\xspace}

\newcommand{\captioner}[0]{\psi\xspace} % 
\newcommand{\textencoder}[0]{f_{T}\xspace}

\newcommand{\absrep}[0]{h_{s}^{A}\xspace}
\newcommand{\conrep}[0]{h_{s}^{C}\xspace}

\definecolor{ConcreteBlue}{RGB}{218,232,252}
\definecolor{AbstractRed}{RGB}{248,206,204}
\definecolor{MethodGreen}{RGB}{213,232,212}

\definecolor{AbstractRed}{HTML}{fecdc0}
\newcommand{\inlineColorbox}[2]{\begingroup\setlength{\fboxsep}{1pt}\colorbox{#1}{\hspace*{2pt}\vphantom{Ay}#2\hspace*{2pt}}\endgroup}

\usepackage[most]{tcolorbox}
\definecolor{block-gray}{gray}{0.96}
\newtcolorbox{myquote}{colback=block-gray,grow to right by=-2mm,grow to left by=-2mm,
boxrule=0pt,boxsep=0pt}

\usepackage[symbol]{footmisc}

%% file: sec/0_abstract.tex
\begin{abstract}
Natural language goes beyond dryly describing visual content. It contains rich abstract concepts to express feeling, creativity and properties that cannot be directly perceived. Yet, current research in Vision Language Models (VLMs) has not shed light on abstract-oriented language.
Our research breaks new ground by uncovering its
wide presence and under-estimated value, with extensive 
analysis.
Particularly, we focus our investigation on the fashion domain, a highly-representative field with abstract expressions. 
By analyzing recent large-scale multimodal fashion datasets, we find that abstract terms have a dominant presence, rivaling the concrete ones, providing novel information, and being useful in the retrieval task. 
However, a critical challenge emerges: current general-purpose or fashion-specific VLMs are pre-trained with databases that lack sufficient abstract words in their text corpora, thus hindering their ability to effectively represent abstract-oriented language. 
We propose a \textit{training-free} and \textit{model-agnostic} method, \methodfull{} (\methodshort), to shift abstract representations towards well-represented concrete ones in the VLM latent space, using pre-trained models and existing multimodal databases.
On the text-to-image retrieval task, despite being training-free, \methodshort{} outperforms the fine-tuned VLMs in both same- and cross-dataset settings, exhibiting its effectiveness with a strong generalization capability. Moreover, the improvement introduced by \methodshort{} is consistent with various VLMs, making it a plug-and-play solution.
%Our code will be publicly available.
\end{abstract}

\iffalse
Natural language goes beyond simple descriptions, encompassing rich abstract concepts that express emotions, creativity, and imperceptible properties. Yet, Vision Language Models (VLMs) have largely overlooked abstract-oriented language.

Our research addresses this gap, focusing on the fashion domain, a representative field rich in abstract expressions. By analyzing large-scale multimodal fashion datasets, we find that abstract terms rival concrete ones in prevalence, provide novel insights, and enhance retrieval tasks. However, most VLMs are pre-trained on datasets with insufficient abstract terms, limiting their ability to represent abstract-oriented language effectively.

To address this, we propose \methodfull{} (\methodshort), a \textit{training-free} and \textit{model-agnostic} method that improves abstract representations in VLM textual embeddings using pre-trained models and existing multimodal datasets. The main idea is to shift abstract representation towards well represented concrete ones in the VLM textual embeddings, exploiting   On text-to-image retrieval tasks, \methodshort{} outperforms fine-tuned VLMs in both same- and cross-dataset settings, demonstrating strong generalization and compatibility across various VLMs as a plug-and-play solution.

Our code will be publicly available."
\fi

%% file: sec/1_intro.tex
\section{Introduction}

Human language extends beyond the straightforward depiction of visual elements and physical attributes presented in images. Evolving over millennia, humans are skilled in expressing complex feelings, spiritual qualities and stimulating inspirations through a nuanced interplay of \textit{abstract} and \textit{concrete} concepts.
In the lexicon literature, \textit{concrete} and \textit{abstract} words are often distinguished based on whether they can be perceived and acted upon~\cite{vigliocco2014neural,brysbaert2014concreteness}.
Such language is extensively applied in fields where creativity and aesthetics are most valued. 
As one of the most representative domains, fashion language, spoken by experts, critics, and general consumers, utilizes terms like ``different", ``sexy", or ``airy", to encapsulate the essence of fashion items, offering richer insights than mere visual descriptions (as demonstrated in Fig.~\ref{fig:teaser}). They are useful to evoke emotions or imagery without specifying perceivable attributes, while concrete attributes like ``strapless'', ``loose'' or ``black'' could potentially limit expressivity across diverse tastes.

This work focuses on the study of abstract-oriented language, specifically in the context of fashion, which is well-represented in large-scale multimodal fashion datasets such as FACAD~\cite{yang2020fashion} and DeepFashion~\cite{liu2016deepfashion}. 
We investigate the usage of adjectives in fashion descriptions, as they often express the item attributes. Our statistical analysis suggests that fashion description is naturally \textit{abstract-oriented}, \ie the presence of abstract adjectives is equal to or larger than the concrete ones. Through our dedicated experiments, we observe that abstract attributes are useful in the retrieval task since they convey novel and complementary information with respect to concrete ones. 
Being more discriminative, abstract-oriented descriptions can effectively enhance retrieval quality and improve retrieval precision.
Furthermore, being able to leverage abstract-oriented descriptions has practical impacts, as users can interact with visual fashion content in human-friendly language at free will. For example, an e-commerce customer could search for items such as ``an oversized summer-mood t-shirt" or ``a sporty top with a fresh collar".

Yet, we discover that such abstract descriptions are under-represented by existing web-scale pre-trained Vision Language Models (VLMs) \cite{radford2021learning}, including the fashion-specific variants~\cite{chia2022contrastive,cartella2023open}, as demonstrated by their poor text-to-image retrieval performance when facing abstract rich natural queries (shown in Fig.~\ref{fig:usefulness-vlmbias}-right). This might be attributed to the properties of the pre-training dataset~\cite{schuhmann2021laion,gadre2024datacomp} crawled from the Internet: our analysis on LAION-400M~\citep{schuhmann2021laion} shows that its textual description is concrete-oriented, with a dominant presence of concrete adjectives over abstract ones. 
Can fine-tuning existing VLMs with multimodal fashion datasets be an effective rescue? Unfortunately, fine-tuning to align visual content with the vast abstract vocabulary hinders the model generalization, as the scale of available fashion datasets constrains it. 
Creating abstract-oriented multimodal databases at million-scale is impractical, as many such creative contents are proprietary or IP-protected. The largest open-source multimodal fashion dataset with natural language descriptions, FACAD~\cite{yang2020fashion}, only contains about 100k items, a few magnitudes less than LAION~\citep{schuhmann2021laion}.

With this data scarcity challenge in mind, we present a novel training-free method, \textit{\methodfull~(\ours)}, that effectively shifts the under-represented abstract-oriented language towards a concrete-oriented one within the representation space of existing VLMs (as demonstrated in Fig.~\ref{fig:teaser}).
\ours~operates in two phases: a one-time \textit{preparation phase} and an \textit{inference phase}. 
The \textit{preparation phase} aims to discover the representation differences between abstract-oriented and concrete-oriented descriptions. To do so, we first \textit{construct a paired Abstract-Concrete multimodal database} by identifying a multimodal dataset with abstract-oriented language and augmenting it with concrete-oriented descriptions produced by image captioning models. Then, we \textit{extract the major representation differences} between the paired descriptions, characterizing the shift with a dimensionality reduction strategy. 
During the \textit{inference phase}, we first utilize a Large Language Model (LLM) to rephrase the abstract-oriented description towards a more concrete version. LLM-based rewriting, however, is still subject to the presence of abstract words. Thus, we further 
augment the textual representation following the shift extracted from the multimodal database.
When evaluated on the text-to-image retrieval task in same- and cross-dataset settings, \ours outperforms even fine-tuned general-purpose and fashion-specific VLMs while being training-free. \ours~also demonstrates model-agnostic effectiveness when combined with various VLMs.

\noindent{\textbf{Our contributions}} are summarized as below:
%\mc{is this way of capitalize the contribution something usual? I don't like it very much, perhaps standard bold will suffice? to discuss}
\begin{itemize}
\item We investigate existing multimodal fashion datasets and prove the \textbf{prevalent presence} of abstract words in the fashion language.
\item We devise indicative experiments to analyze the textual content, and prove that abstract attributes convey \textbf{novel information} and are \textbf {useful in fashion retrieval}.
\item We present the problem that existing VLMs \textbf{under-represent} abstract-oriented descriptions, with an empirical verification in the task of text-to-image retrieval.    
\item We propose an \textbf{effective training-free method \ours{}} to bridge the abstract-concrete representation shift, outperforming zero-shot VLMs by up to
$+12.6\%$, and fine-tuned ones by $+2.0\%$ in the H@1 metric.  
\end{itemize}

%% file: sec/2_related.tex
\section{Related Work}
\label{sec:related}

\boldedparagraph{Fashion datasets and text descriptions.}
Fashion image datasets saw a shift in textual descriptions starting in the mid-2010s.
Early datasets like FashionStyle~\citep{TakagiICCVW2017} focused on labeling entire outfits with broad styles (``goth'', ``retro''). Others like~\citep{SimoSerraCVPR2015,InoueICCVW2017} used noisy tags from a limited vocabulary. Subsequently, datasets like iMaterialist~\citep{guo2019imaterialist}, FashionMNIST~\citep{xiao2017fashion}, Fashion200K~\citep{han2017automatic} adopted more fine-grained text descriptions, including attributes, class labels, and multiple weighted tags. Over the past few years, the trend has been to employ natural language descriptions, with human-annotated captions being considered in datasets such as Fashion-Gen~\citep{rostamzadeh2018fashion}, Fashion IQ~\citep{wu2021fashion}, and KAGL~\citep{param_aggarwal_2019}. However, these descriptions are aimed towards image generation~\citep{rostamzadeh2018fashion} or pair-wise comparison tasks~\citep{wu2021fashion}, and mainly capture fine-grained concrete properties, with retrieval not being their primary goal. Consequently, textual representations focus on visually grounded properties, namely, colors, fabrics used, and the presence of specific details or accessories. 
Coarser information about the garments, usually associated with abstract properties such as emotive qualities (\eg ``aggressive'', ``minimal''), is typically overlooked. 
More recently, fashion applications have taken advantage of aligned text-image representations and have started considering more complex descriptions~\citep{yang2020fashion, liu2016deepfashion}. 
In DeepFashion~\citep{liu2016deepfashion}, data samples are crawled from fashion e-commerce platforms. It contains descriptions involving a range of fashion trends, moods, and 
abstract attributes of various garment items. 
Similar approach characterize FACAD~\citep{yang2020fashion}, while it features a larger scale and generally more compact descriptions.

\boldedparagraph{Retrieval in the Fashion domain.} 
Early fashion retrieval systems were initially used for matching real-world garments to shop advertisements~\citep{hadi2015buy, liu2016deepfashion, huang2015cross, ji2017cross, wang2011clothes, di2013style}. In more recent literature, some systems rely on feedback-based techniques, where users provide an image and text describing desired changes, and the system retrieves similar images reflecting those modifications~\citep{goenka2022fashionvlp, zhuge2021kaleido, anwaar2021compositional, lee2021cosmo, baldrati2022effective}.
Currently, Vision-Language Models (VLMs) like CLIP~\citep{radford2021learning} have revolutionized retrieval tasks with their strong generalization power. These models leverage web-scale text-image data through contrastive learning.
Recent research builds on CLIP-like models for text-only retrieval, adapting general-purpose models to the fashion domain~\citep{cartella2023open, chia2022contrastive}. Starting from a pre-trained model on large-scale data~\citep{schuhmann2021laion}, F-CLIP~\citep{chia2022contrastive} adapts to the specialized domain by employing proprietary data. Nonetheless, from the qualitative inspection available in their publication, the textual descriptions in the training data mainly contain concrete attributes. Similarly, OF-CLIP~\citep{cartella2023open} focuses on open-source data for fine-tuning to the fashion domain. However, these data only involve concrete-oriented descriptions, such as FASHION200K~\citep{han2017automatic}, iMATERIALIST~\citep{guo2019imaterialist}, Fashion Gen~\citep{rostamzadeh2018fashion} and Fashion IQ~\citep{wu2021fashion}.

%% file: sec/3_preliminary.tex
\section{Analyzing the Language of Fashion}
\label{sec:preliminary}

In this section, we analyze recent large-scale multimodal datasets in the fashion domain, and highlight four main findings:
i) abstract attributes are abundantly used in the textual descriptions of fashion images; ii) abstract attributes convey novel information; iii) abstract attributes could boost retrieval performance; iv) current VLMs under-represent abstract-oriented language. %This section motivates the experiments of Sec.~\ref{sec:experiments}.

\begin{figure}[t!]
    \centering
    \includegraphics[width=\linewidth]{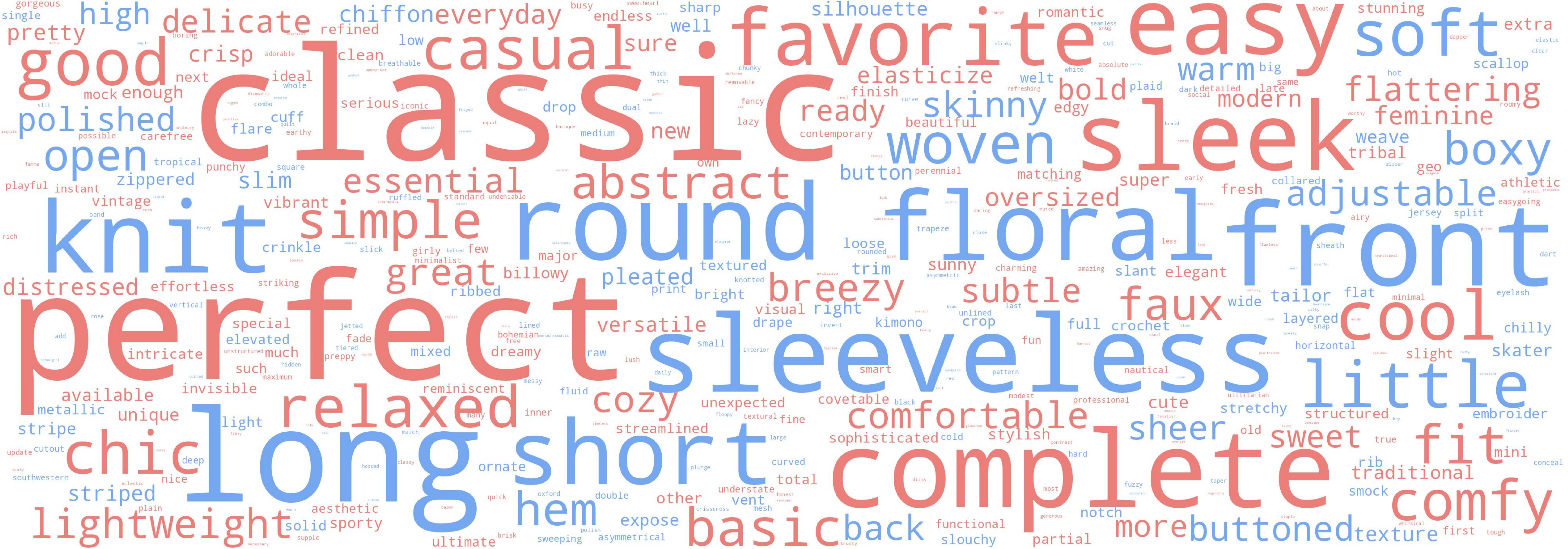}
    % \caption{Wordcloud of the abstract (in orange) and concrete (in blue) adjectives in the DeepFashion dataset. The larger font represents a higher frequency.}
    % \label{fig:word_clouds}
    \includegraphics[width=0.95\linewidth]{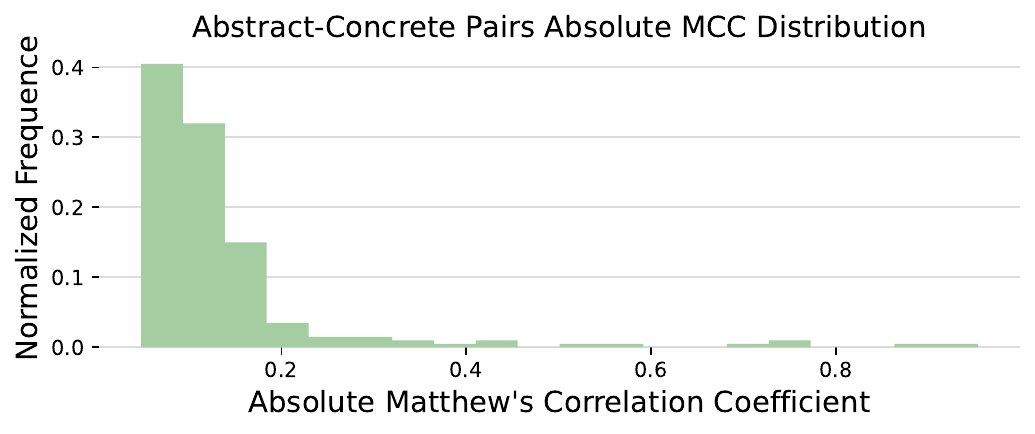}
    \caption{\textbf{Top:} Wordcloud of the \inlineColorbox{AbstractRed}{abstract} and \inlineColorbox{ConcreteBlue}{concrete} adjectives in the DeepFashion dataset. The larger font represents a higher frequency. \textbf{Bottom:} Distribution of maximum absolute Matthews's Correlation Coefficient (MCC) between each abstract attribute and concrete ones in DeepFashion.
    % eval split. 
    % MCC close to 0 indicates no correlation between two attributes while MCC = 1 (-1) suggests a positive (negative) correlation. 
    The peak near 0 reveals that the majority of abstract attributes have a low correlation with the concrete ones.}\label{fig:wordcloud-mcc}
\end{figure}

\noindent\textbf{\textit{Datasets.}} We study DeepFashion~\citep{liu2016deepfashion} and FACAD~\citep{yang2020fashion} as popular multimodal fashion 
 datasets, and LAION~400M~\citep{schuhmann2021laion} as one of the main multimodal datasets used to pre-train large VLMs~\citep{cherti2023reproducible, sun2023eva}. 
Specifically, both \textit{DeepFashion}~\citep{liu2016deepfashion} and \textit{FACAD}~\citep{yang2020fashion} contain data collected from real fashion e-stores. The items cover different garment classes, from pants, shirts, and skirts to shoes and jewelry. Each item is paired with a multi-sentence description (median 37 words) originally used by the sellers on the platform, reflecting the exact vocabulary used in the fashion domain.
On the contrary, \textit{LAION~400M}~\citep{schuhmann2021laion} is composed of web-crawled images, along with short descriptions, usually coming from the image metadata.

\noindent\textit{\textbf{Attribute extraction and categorization.}}
We focus on \textit{attributes} that describe certain property of an item, composed of an adjective with an associated noun (if present) (\eg ``relaxed neckline'').
We develop an attribute extraction and categorization pipeline that leverages spaCy~\citep{spacy}, a popular NLP tool, to localize the attributes present in the item descriptions. Finally, we categorize them into abstract and concrete based on the adjectives using a well-established \emph{concreteness lexicon}~\cite{brysbaert2014concreteness}. %containing the average perception of words in terms of concreteness (rating between 1 and 5). 
We classify attributes to be abstract if the ratings of their adjectives are below $3.0$, following the threshold strategy of~\cite{brysbaert2014concreteness}. We refer the reader to the \suppmat{} for more details on the extraction and categorization pipeline.

\noindent\textbf{i) Fashion descriptions are abstract-oriented.}\label{find1}
Tab.~\ref{tab:adj_freq} reports the number of occurrences of adjectives in different datasets. The ratio 
between total abstract and concrete
adjectives differ greatly among DeepFashion, FACAD, and LAION~400M. In FACAD and DeepFashion, abstract adjectives are as frequent (or more frequent) than concrete adjectives, while descriptions in LAION~400M are heavily skewed towards concrete adjectives. 

Fig.~\ref{fig:wordcloud-mcc}-top presents the cloud of words of the extracted concrete and abstract adjectives of DeepFashion. 
The most frequent concrete adjectives are ``long'', ``sleeveless'', ``front'', and  ``round'', and all of them are clearly effective in describing the fashion items.   
Nonetheless, fashion language isn't about objectively describing a product, it also expresses feelings, reflects preferences, and unveils desires, \ie stylistic nuances that all have a very specific translation in terms of appearance.
The cloud of words exhibits exactly this function of the fashion language.  
Abstract adjectives like ``classic'', ``sleek'', ``casual'' paint a picture in the reader's mind of the feeling the item conveys, while others like ``relaxed'' or ``comfy'' can be used to focus on the details of the silhouette, \eg ``a comfy neckline''.

\begin{table}
  \resizebox{\linewidth}{!}{%
    \begin{tabular}{l ccc ccc}
        \toprule
        \multirow{2}{*}{\textbf{Dataset}}& \multicolumn{3}{c}{\textbf{Concrete occurrences}} 
            &\multicolumn{3}{c}{\textbf{Abstract occurrences}}\\ 
        \cmidrule(lr){2-4}
        \cmidrule(lr){5-7}
         & Unique & Total & Per desc. & Unique & Total & Per desc.\\
        \midrule 
        DeepFashion & 625 & 10,905 & 3 & 716 & 14,045 & 3\\
        FACAD & 2795 & 202,883 & 2 & 2019 & 194,528 & 2\\
        LAION~400M & 14,217 & 26M & 0 & 12,844 & 19M & 0\\

        \bottomrule
    \end{tabular}
    }
    \caption{Occurrences for each adjective in different datasets. In the ``Per description'' column (Per desc.), the median is reported. It is clear to see how, in abstract-oriented datasets, abstract adjectives are as common (if not more) as concrete ones. In concrete-oriented datasets (100M subset of LAION~400M), abstract adjectives are the minority.}
  \label{tab:adj_freq}
\end{table}

\noindent\textbf{ii) Abstract attributes convey novel information.}
We aim to demonstrate that the abstract and concrete attributes carry different information when describing fashion items. Intuitively, if an abstract attribute always co-occurs with a concrete one, then they are highly correlated and convey similar information. Instead, if the abstract attribute is rarely bound to other concrete ones, then it conveys diverse information. 
We thus conduct a correlation study, showing a low correlation between abstract and concrete attributes. 

We limit our analysis to the set of 200 most frequent concrete $\mathcal{A}^{c}$ and abstract $\mathcal{A}^{a}$ attributes, resulting in $K=400$ attributes considered in the correlation study. Leveraging the Matthew Correlation Coefficient (MCC)~\citep{matthews1975comparison}, we define, for each abstract attribute $\mathcal{A}^{a}_i$, its correlation $\Phi_{i}$ against concrete attributes as the maximum absolute correlation $\phi$ between the abstract attribute $i$ and all the concrete ones.
\begin{equation}
\Phi_{i} = \max_{\mathcal{A}^c_j\in \mathcal{A}^{c}} \left|\phi(\mathcal{A}^{a}_{i}, \mathcal{A}^{c}_{j})\right|,
\end{equation}

Fig.~\ref{fig:wordcloud-mcc}-bottom shows the distribution of the maximum absolute MCC $\Phi$ of $\mathcal{A}^{a}$ on DeepFashion. The distribution is highly skewed towards left, \ie~most abstract attributes have a low correlation to the concrete ones. A small set of abstract-concrete pairs exists with a high correlation. By manual inspection, we notice that those pairs often refer to specific attributes jointly used. For instance, (``faux belt'', ``leather belt'') is common because of ``faux leather belt'', and similarly for ``invisible side zipper''.

\noindent\textbf{iii) Abstract attributes are useful for fashion retrieval.}
We investigate whether, given the same number of attributes in the text query, the description containing abstract attributes is more discriminative than that of those containing concrete ones. To limit ambiguity induced by text/image representation in the retrieval performance, we perform the retrieval task with an oracle retrieval system that is purely based on attribute matching.      

Given the set $D$ of descriptions that contain at least one of the $K$ investigated attributes, we build the matrix $P$, where the rows are the descriptions $d \in D$, and the $K$ columns report the binary presence of each of the $K$ attributes in that description.
An ideal retrieval system, once queried with a set of attributes $q$, will retrieve a set of items $R_q$ where each item contains all (but not only) the attributes present in $q$.
Each query will be a set of attributes present in the matrix $P$,  meaning that we use each item attribute set to query the matrix itself.
Let $N_q = |R_q|$ be the number of retrieved items when querying with attributes indexed by $q$.
We can quantify how discriminative the query is by 
computing the retrieval precision $p_q = 1/N_q$, with a $p_q = 1$ indicating the query 
contains a \textit{unique} set of attributes.

\begin{figure}
    \centering
    \begin{subfigure}[b]{.49\linewidth}
    \centering
    \includegraphics[width=\linewidth]{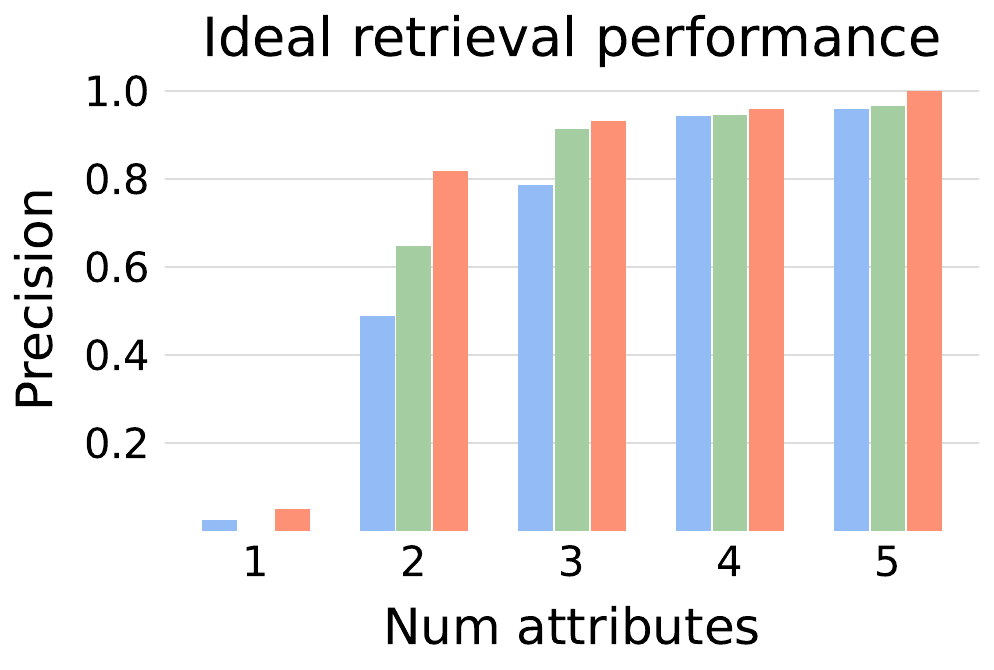}
    \end{subfigure}%
    \hspace{-5pt}
    \begin{subfigure}[b]{.49\linewidth}
    \centering
    \includegraphics[width=\linewidth]{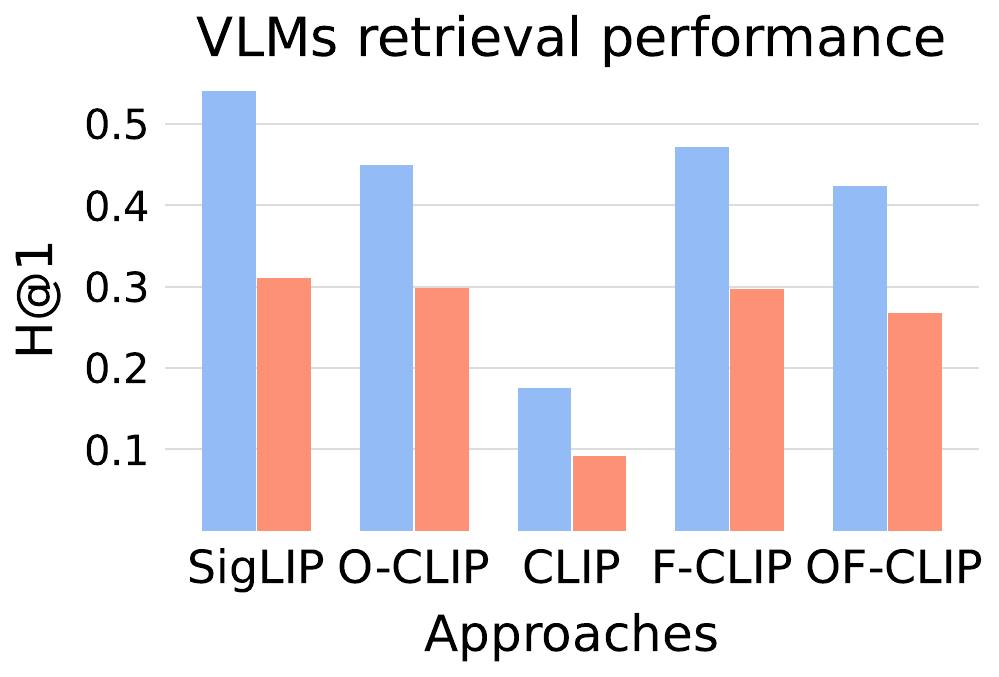}
    \end{subfigure}
    \caption{\textbf{Left:} Retrieval performance of an ideal system on Deepfashion original descriptions when the majority of present attributes are \inlineColorbox{ConcreteBlue}{concrete},  \inlineColorbox{AbstractRed}{abstract} or \inlineColorbox{MethodGreen}{mixed}. Abstract attributes allow for better retrieval performance. \textbf{Right:} performance of current VLMs on DeepFashion when using original \inlineColorbox{AbstractRed}{abstract} descriptions or \inlineColorbox{ConcreteBlue}{concrete} VLM generated ones. Current VLMs achieve better performance with concrete-oriented descriptions.
    \label{fig:usefulness-vlmbias}}
\end{figure}

Fig.~\ref{fig:usefulness-vlmbias}-left reports the average precision of sentences when grouping them by the number of attributes in the sentence and categorizing them based on the most present attribute categories 
(where mixed indicates the ratio between the number of abstract and concrete attributes is between $0.33$ and $0.66$).
We consider sentences containing at most 5 attributes, consisting of $\sim90\%$ of the data. 
The figure shows that abstract attributes consistently provide more or equal information to concrete ones, as highlighted by the higher precision across a varying number of attributes in both mixed and abstract descriptions.
In particular, with smaller numbers of attributes, \ie one-three, we observe that abstract attributes can achieve much higher precision than concrete attributes. This suggests that abstract attributes encapsulate more specialized information than concrete ones, allowing for more faithful descriptions with fewer words.

\noindent\textbf{iv) Current VLMs under-represent abstract language.} 
We probe the capability of different state-of-the-art VLMs to represent and handle abstract-oriented descriptions.
In the first step, we generate concrete descriptions from visual data using state-of-the-art captioning models. As highlighted by the large presence of concrete adjectives in Tab.~\ref{tab:vlm_stats}, generated descriptions mainly rely on concrete attributes to convey visual information. In the second step, we evaluate the retrieval performance of a model when queried with original abstract descriptions  \emph{vs} concrete-oriented ones. We align with the retrieval literature~\citep{chia2022contrastive, cartella2023open} and use Hit-Rate@1, which considers the retrieval successful if the image most similar to the description is correct. From Fig.~\ref{fig:usefulness-vlmbias}-right, it is clear how both general-purpose and fashion-specialized VLMs (F-CLIP, OF-CLIP) strongly favor concrete descriptions over abstract ones, suggesting an inability to properly encode abstract attributes.

\begin{table}
  \resizebox{\linewidth}{!}{%
    \begin{tabular}{l ccc ccc}
        \toprule
        \multirow{2}{*}{\textbf{Model}}& \multicolumn{3}{c}{\textbf{Concrete occurrences}} 
        &\multicolumn{3}{c}{\textbf{Abstract occurrences}}\\ 
        \cmidrule(lr){2-4}
        \cmidrule(lr){5-7}
         & Unique & Total & Per desc. & Unique & Total & Per desc.\\
        \midrule
        \multicolumn{7}{c}{\cellcolor{Gray!16}\textbf{Captioning Models}}\\
        Qwen2-VL & 245 & 12,819 & 3 & 98 & 2,623 & 0   \\
        CogVLM2 & 351 & 19,042 & 5 & 265 & 10,774 & 2    \\
        \multicolumn{7}{c}{\cellcolor{Gray!16}\textbf{Large Language Models}}\\
        Phi3 & 425 & 7,737 & 2 & 281 & 3,742 & 1    \\
        LLaMa 3.1 8B & 405 & 7,778 & 2 & 199 & 2,675 & 0    \\
        \bottomrule
    \end{tabular}
    }
    \caption{The number of occurrences for each adjective category in the captioned train split of DeepFashion. As we can see, both captioning models and LLMs are biased towards concrete properties.}
  \label{tab:vlm_stats}
\end{table}

%% file: sec/4_method.tex
\section{\methodfull{}}
\label{sec:method}
\begin{figure*}[t!]
    \centering
    \includegraphics[width=0.98\linewidth]{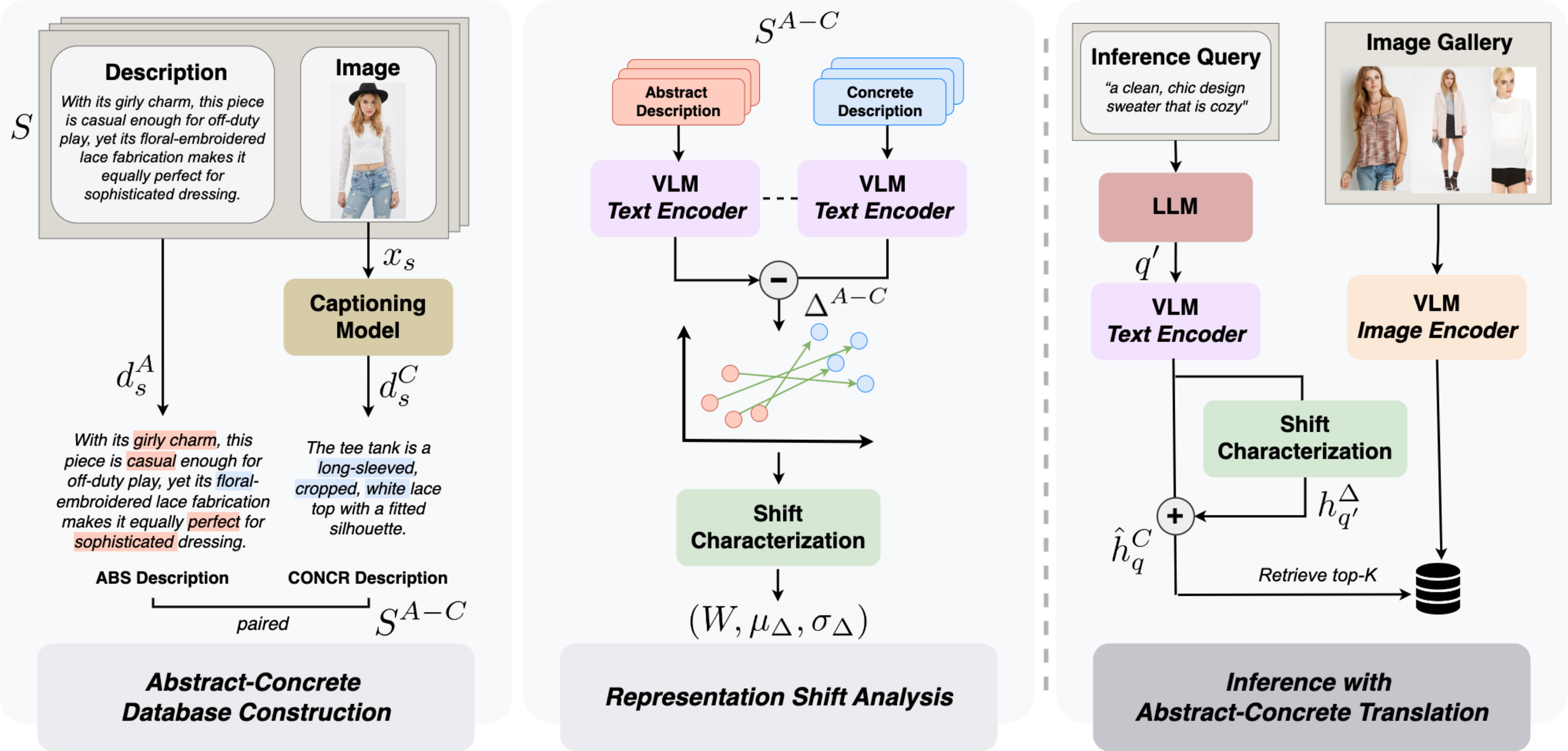}
    \caption{Overview of our two-phase \methodfull{} (\methodshort). During the \textit{preparation phase}, \methodshort~conducts a first \textit{database construction step}, processing an Abstract-Concrete (\AC) database by using an image captioning model to produce concrete-oriented captions describing the images. Then, in the \textit{representation shift analysis step}, \methodshort{} analyzes the main representation shifts among the paired \AC descriptions with a dimensionality reduction strategy. During the \textit{inference phase}, \methodshort~first prompts a frozen LLM to rephrase the abstract-oriented description to convert the abstract-oriented language into a more concrete-oriented expression. Then, \methodshort~enhances the VLM textual representation by compensating with the main shifts extracted from the \AC multimodal database. This allows \methodshort~to perform better on downstream multimodal tasks with abstract-oriented language, \eg text-to-image retrieval, without any training.}
    % \caption{Overview of our two-phase \methodfull{} (\methodshort). In the first \textit{preparation phase},  the \textit{database construction step}, \methodshort~processes an Abstract-Concrete (\AC) database, using an image captioning model to produce concrete-oriented captions describing the images. In the second \textit{representation shift analysis phase}, \methodshort{} analyzes the main representation shifts among the paired \AC descriptions with a dimensionality reduction strategy. Finally, at the \textit{inference phase}, we first prompt a frozen LLM to rephrase the abstract-oriented description, an important step to convert the abstract-oriented language into more concrete-oriented expression. We then enhance the textual representation by compensating with the main differences extracted from \AC multimodal database, to perform better on downstream multimodal task with abstract-oriented language, \eg text-to-image retrieval, without training existing VLMs.}
    \label{fig:method}
\end{figure*}

The proposed \methodfull{}~(\methodshort) augments the latent representation of abstract-oriented descriptions to process fashion items textual data more effectively. Our approach is guided by the intuition that aligning these representations closer to the well-modeled concrete representations within the VLM latent space can enhance the model's performance on downstream tasks. As illustrated in Fig.~\ref{fig:method}, the method comprises two phases. During the \textit{first preparation phase}, the approach characterizes the shift from abstract to concrete representations. To this end, the \textit{database construction step} constructs an abstract-concrete (\AC) database from existing abstract-oriented multimodal data: we leverage an image captioning model to produce captions describing the images, providing paired descriptions of the same visual data. As shown in Tab.~\ref{tab:vlm_stats}, such generated captions are more visually concrete-oriented.
Later on, the \textit{ representation shift analysis step} leverages a dimensionality reduction strategy to identify the main representation shift among the paired \AC descriptions. 
These shift directions capture the discrepancies in the latent space between concrete-oriented and abstract-oriented descriptions.
Once the shift has been characterized, the later \textit{second inference phase} can leverage it to augment all provided queries.
Here, we start by utilizing a frozen LLM to rephrase the abstract-oriented description, an important step to convert the abstract-oriented language into a more concrete-oriented expression. Then, we further enhance the textual representation by shifting it towards concrete ones along the dimensions extracted from the \AC multimodal database.

\noindent\textbf{\AC Database Construction.}
Starting from a multimodal database in the fashion domain containing abstract-oriented textual descriptions, we leverage a pre-trained image captioning model to construct the Abstract-Concrete (\AC) database, aiming to pair abstract and concrete descriptions associated by their images (the left block of Fig.~\ref{fig:method}).

Consider a multimodal database $S=\{(d^A_s, \mathcal{I}_s)\}_s$ where each clothing item $s$ has an abstract-oriented description $d^A_s$ and a set of corresponding images $\mathcal{I}_s$. 
For each item, we consider the visual sample $x_s \in \mathcal {I}_s$, and use the frozen captioning model, denoted as $\captioner$, to output a concrete-oriented description $d^C_s$:
\begin{equation}
    d^C_s = \captioner(x_s, p_v),
\end{equation}
where $p_v$ is the text prompt given to the captioning model for describing the clothing item $s$ (see \suppmat{} for prompt details.). %visually grounded properties like shape, pattern and garment details.

Formally, we construct our \AC database $S^{\text{\AC}}$ as:
\begin{equation}
S^{\text{\AC}} = \{(d^A_s, d_s^C) \mid s \in S\}.
\end{equation}

\noindent\textbf{\AC Representation Shift Analysis.}
The paired abstract-concrete descriptions allow us to analyze the main differences between the textual representations of abstract-oriented and concrete-oriented descriptions anchored by the same visual content (the middle block of Fig.~\ref{fig:method}).   

For each item $s$, we use the text encoder of a VLM $\textencoder$ to encode its abstract-oriented and concrete-oriented descriptions ($d^A_s$ and $d^C_s$), obtaining:
\begin{align}
\centering
    \absrep &= \textencoder(d^A_s),\quad s \in S,\\\nonumber
    \conrep &= \textencoder(d^C_s),\quad s \in S.
\end{align}

Let $H^A$ and $H^C$ denote the vectorized abstract-oriented and concrete-oriented representations of all items in the database, respectively, \ie the $s$-th row of $H^A$ is $\absrep$, and the $s$-th row of $H^C$ is $\conrep$.
We aim to identify the dominant differences among $H^A$ and $H^C$. To this end, we perform subtraction between $H^C$ and $H^A$ as:
\begin{equation}
\Delta^{\text{\AC}} = H^C - H^A. 
\end{equation}

We then standardize $\Delta^\text{\AC}$ to have zero-mean and unit variance by rescaling using statistics $\mu_\Delta, \sigma_\Delta$ denoting the shift mean and standard deviation, respectively. 
To extract the main directions that best capture the differences between $H^C$ and $H^A$, we apply a Principal Component Analysis (PCA)~\citep{shlens2014tutorial}:
\begin{equation}
    W = \text{PCA}(\Delta^\text{\AC}, k),
    \label{eq:pca}
\end{equation}
obtaining the shift projector $W \in \mathbb{R}^{l \times k}$, where $l$ is the dimension of the textual embeddings and $k$ is the number of considered components. 
$W$ models the main directions along which lies the information that $H^A$ struggles to represent with respect to the concrete counterpart $H^C$.

\noindent\textbf{\ours at inference.}
\methodshort converts an abstract-oriented textual representation to a more concrete-oriented one, in order to enhance the downstream tasks that require multimodal alignment (the right block in Fig.~\ref{fig:method}). 
In the context of text-to-image retrieval task, given an abstract-oriented text query $q^A$, we first leverage a LLM, denoted by $g$, to rephrase it to a more concrete version $q'$:
\begin{equation}
    q' = g(q^A, p_r),
\end{equation}
where $p_r$ is the rephrasing prompt. 

We then embed $q'$ with the VLM text encoder $\textencoder$, obtaining its representation $h_{q'}$. As demonstrated in Tab.~\ref{tab:vlm_stats}, while the LLM can improve language concreteness, the rephrased text $q'$ may still contain abstract words. Thus, we further leverage the characterization of the representation shift $(W, \mu_\Delta, \sigma_\Delta) $ obtained from the previous step to shift the representation of $h_{q'}$ towards a more concrete-oriented representation $\hat{h}_q^C$. 
Specifically, we compute $h_{q'}^\Delta$ from $h_{q'}$ through the representation shift characterization, as expressed below:
\begin{align}
\centering
h^\Delta_{q'} &= N(h_{q'})\, WW^T * \sigma_\Delta + \mu_\Delta, \\\nonumber
\hat{h}_q^C &= h_{q'} + h^\Delta_{q'},
\end{align}
where $N(\cdot)$ indicates the standardization operation using the mean and standard deviation among query representations. After the projection, the representation is then re-scaled with the $\Delta^{\AC}$ statistics ($\mu_\Delta$, $\sigma_\Delta$). Intuitively, the projection into the main components along $W$ extracts the meaningful differences between the abstract-oriented and concrete-oriented textual representations. Hence, re-scaling using $\Delta^{\AC}$ statistics projects the representation back into the VLM manifold, allowing the downstream cross-modal task.  
The final 
$\hat{h}_q^C$ combines the representation of the rephrased text $h_{q'}$ and the representation shift. 
Notice that all queries are augmented during the inference phase using the $(W, \mu_\Delta, \sigma_\Delta)$ that characterize the major variations from abstract to concrete representations. Due to $(W, \mu_\Delta, \sigma_\Delta)$ being pre-processed during the preparation phase, the \AC database is not required in the inference phase.

%% file: sec/5_experiments.tex
\section{Experiments}
\label{sec:experiments}
The experiments focus on the text-to-image retrieval task, serving as a practical scenario to demonstrate how VLMs can benefit from \methodshort.  
Here, we first introduce the evaluation protocol and implementation details. Then, we discuss the main comparison conducted on DeepFashion~\cite{liu2016deepfashion}, together with ablation studies to prove the benefit of the Language Rewriting and Representation Shift modules.

\noindent\textbf{Retrieval task.}
We follow common practice~\citep{chia2022contrastive, cartella2023open, radford2021learning} and simulate the user querying the $q$-th item by using the textual description $d_q$, and expect the system to retrieve the images within the set $\mathcal{I}_q$. Specifically, in the first step, the visual encoder extracts embeddings for all the images in the queried collection, creating a gallery of image embeddings $H_I$. Then, given a query $d_q$, the text encoder extracts embeddings $h_q$, which are used to retrieve the most similar images $\mathcal{\hat{I}}_q$ by selecting the corresponding top-K most similar image embeddings from $H_I$ under cosine similarity. Similarly, given a query $d_q$, \methodshort shifts abstract-oriented queries to obtain the final embedding $\hat{h}^C_q$,  which is used to retrieve the top-K most similar images with the same procedure.

\noindent\textbf{Metrics.} We quantitatively evaluate the retrieval performance using commonly-used metrics~\cite{cartella2023open, chia2022contrastive}, \ie Recall@K (R@K) and Hit-Rate@K (H@K). Recall@K measures the percentage of all relevant images corresponding to item $q$ presented in the $K$ retrieved ones $\mathcal{\hat{I}}_q$. Hit-Rate@K evaluates the number of successful queries out of all those performed. Formally, given a description $d_q$, we consider the query a success if at least one of the images in $\mathcal{I}_q$ is present in $\mathcal{\hat{I}}_q$. 
Following the protocol presented in~\citep{cartella2023open}, we report the results at different values of $K$, with $K\in \{1,5,10\}$, and report the average metric over all the queries.

\noindent\textbf{Baselines.} To assess the benefit of \methodshort we consider various families of VLMs as retrieval models, including SigLIP~\citep{zhai2023sigmoid}, CLIP~\citep{radford2021learning}, O-CLIP~\citep{cherti2023reproducible} and EVA-CLIP~\citep{sun2023eva}. In particular, for each family of retrieval models, we consider a zero-shot evaluation of the general purpose model to the retrieval task. We further compare with their fine-tuned version to available abstract-oriented data, denoted as \texttt{<model>-ft-<data>}. As representative of state-of-the-art approaches, we compare with fashion-specialized retrieval approaches, namely F-CLIP~\citep{chia2022contrastive} and OF-CLIP~\citep{cartella2023open}. Finally, we also evaluate the change in performance when scaling to a larger number of parameters. 

\noindent\textbf{Implementation Details.}
As \methodshort is architecture-agnostic, we integrate the proposed strategy across several CLIP-like pre-trained models. For primary evaluations, we use \texttt{SigLIP~(ViT-B-16)}~\citep{zhai2023sigmoid} as the retrieval model unless stated otherwise. The \AC database builds upon an Int4-quantized \texttt{Qwen2-VL-7B}~\citep{wang2024qwen2} as the captioning model, prompted to generate captions that focus on the fashion item of interest, specifically by querying its item class. The representation shift analysis step retains the top $k=600$ principal components, which is an empirical choice. During inference, \methodshort utilizes \texttt{Llama-3.1-8B} to generate an initial concrete version of the original query. See \suppmat

\subsection{Results}
\begin{table}[t!]
\footnotesize
    \centering
    \resizebox{\linewidth}{!}{    
    \begin{tabular}{l ccc ccc ccc}
        \toprule
        \textbf{Model} & \textbf{R@1} & \textbf{R@5} &\textbf{R@10} & \textbf{H@1} & \textbf{H@5} &\textbf{H@10} \\ 
        \midrule
        \multicolumn{7}{c}{\cellcolor{Gray!16}\textbf{Zero-Shot}}\\
    
      \cellcolor{ConcreteBlue}SigLIP         & .062 & .228 & .322 & .311 & .536 & .639 \\
      \cellcolor{ConcreteBlue}CLIP           & .019 & .061 & .091 & .092 & .201 & .278 \\
      \cellcolor{ConcreteBlue}O-CLIP         & .061 & .204 & .291 & .298 & .527 & .637 \\
      \cellcolor{ConcreteBlue}EVA-CLIP       & .045 & .151 & .218 & .225 & .412 & .517 \\
      \cellcolor{ConcreteBlue}F-CLIP         & .060 & .205 & .292 & .297 & .512 & .618 \\
      \cellcolor{ConcreteBlue}OF-CLIP        & .054 & .181 & .260 & .268 & .491 & .606 \\
      \midrule
      \multicolumn{7}{c}{\cellcolor{Gray!16}\textbf{Same-Dataset (DeepFashion $\rightarrow$ DeepFashion)}}\\
      \cellcolor{AbstractRed}SigLIP-ft-df       & \underline{.083} & \textbf{.307} & \textbf{.421} & \underline{.417} & \underline{.659} & \underline{.753} \\
      \cellcolor{AbstractRed}CLIP-ft-df        & .029 & .095 & .142 & .140 & .299 & .393 \\
       \cellcolor{AbstractRed}O-CLIP-ft-df      & .080 & .272 & .378 & .398 & .640 & .740 \\
       \cellcolor{AbstractRed}EVA-CLIP-ft-df    & .063 & .219 & .316 & .317 & .539 & .654 \\
       \cellcolor{AbstractRed}F-CLIP-ft-df    & .074 & .254 & .350 & .363 & .603 & .705 \\
       \cellcolor{AbstractRed}OF-CLIP-ft-df    & .072 & .245 & .344 & .358 & .607 & .712 \\
      \cellcolor{MethodGreen}\textbf{\methodshort-df (Ours)}   & \textbf{.089} & \underline{.303} & \underline{.409} & \textbf{.437} & \textbf{.665} & \textbf{.756} \\
    \midrule
    \multicolumn{7}{c}{\cellcolor{Gray!16}\textbf{Cross-Dataset (FACAD $\rightarrow$ DeepFashion)}}\\
\cellcolor{AbstractRed}SigLIP-ft-facad      & \underline{.070} & \underline{.259} & \underline{.364} & \underline{.352} & \underline{.568} & \underline{.681} \\
        \cellcolor{AbstractRed}CLIP-ft-facad        & .023 & .079 & .122 & .114 & .265 & .358 \\
       \cellcolor{AbstractRed}O-CLIP-ft-facad      & .065 & .230 & .324 & .324 & .562 & .667 \\
       \cellcolor{AbstractRed}EVA-CLIP-ft-facad    & .049 & .169 & .249 & .246 & .457 & .561 \\
       \cellcolor{AbstractRed}F-CLIP-ft-facad    & .060 & .206 & .295 & .295 & .526 & .631 \\
       \cellcolor{AbstractRed}OF-CLIP-ft-facad    & 059 & .209 & .298 & .298 & .542 & .649 \\
        \cellcolor{MethodGreen}\textbf{\methodshort-facad (Ours)}   & \textbf{.087} & \textbf{.302} & \textbf{.407} & \textbf{.428} & \textbf{.661} & \textbf{.754} \\
\bottomrule
    \end{tabular}
    }
    \caption{
       Results on Deepfashion. In \textbf{bold} the best results, while \underline{underlined} are the second best. \inlineColorbox{MethodGreen}{\methodshort} proves to be the best (or second best) method \textit{w.r.t.} ~\inlineColorbox{ConcreteBlue}{zero-shot} and \inlineColorbox{AbstractRed}{fine-tuned} models in both the same-dataset and cross-dataset settings.} \label{tab:results-deepfashion}
\end{table}
\noindent\textbf{Zero-shot + same-dataset retrieval.}
In the first two blocks of Tab.~\ref{tab:results-deepfashion}, we report the results of models evaluated on the DeepFashion evaluation set. 
In the same-dataset setting, the fine-tuned models are trained on the DeepFashion training set, while \ours leverages the DeepFashion training set to construct the \AC database (\texttt{\methodshort-df}). 
Firstly, we note that shifting 
the representations from abstract to concrete allows for a large improvement with respect to the zero-shot approaches (the top block): the result holds for both general purpose models (up to +12.9\% H@5 \textit{w.r.t.} the second-best SigLIP) and fashion-specialized ones (+2.9\% R@1 \textit{w.r.t.} F-CLIP). Interestingly, despite requiring no training, \methodshort proves to be the best or the second best model even when compared to fine-tuned models (second block). 
While the higher precision and recall at $k=1$ shows the ability of the model to select the closest sample, the benefit is especially evident in the H@K metric, where \methodshort achieves the largest gap ($+2\%$ H@1) \textit{w.r.t.} the strongest fine-tuned model.

\begin{table}[t]
    \centering
\resizebox{\linewidth}{!}{
    \begin{tabular}{l c ccc ccc}
        \toprule
        \textbf{Model} & \textbf{Backbone} & \textbf{R@1} & \textbf{R@5} &\textbf{R@10} & \textbf{H@1} & \textbf{H@5} &\textbf{H@10} \\ 
        \midrule
\cellcolor{ConcreteBlue}SigLIP & ViT-B-16 & .062 & .228 & .322 & .310 & .536 & .639 \\
\cellcolor{MethodGreen}\textbf{SigLIP-\methodshort} & ViT-B-16 & \textbf{.089} & \textbf{.303} & \textbf{.409} & \textbf{.437} & \textbf{.665} & \textbf{.756} \\
\hdashline
\cellcolor{ConcreteBlue}SigLIP & ViT-L-16-384 & .094 & .348 & .473 & .458 & .679 & .775 \\
\cellcolor{MethodGreen}\textbf{SigLIP-\methodshort} & ViT-L-16-384 & \textbf{.108} & \textbf{.385} & \textbf{.513} & \textbf{.527} & \textbf{.736} & \textbf{.823} \\
\midrule
\cellcolor{ConcreteBlue}O-CLIP & ViT-B-32 & .061 & .204 & .290 & .298 & .527 & .637 \\
\cellcolor{MethodGreen}\textbf{O-CLIP-\methodshort}& ViT-B-32 & \textbf{.062} & \textbf{.210} & \textbf{.298} & \textbf{.312} & \textbf{.537} & \textbf{.655} \\
\hdashline
\cellcolor{ConcreteBlue}O-CLIP & ViT-L-14 & .085 & .301 & .411 & .412 & .660 & .755 \\
\cellcolor{MethodGreen}\textbf{O-CLIP-\methodshort} & ViT-L-14 & \textbf{.091} & \textbf{.313} & \textbf{.422} & \textbf{.447} & \textbf{.676} & \textbf{.771} \\
\hdashline
\cellcolor{ConcreteBlue}O-CLIP & ViT-H-14 & .091 & .319 & .432 & .447 & .673 & .772 \\
\cellcolor{MethodGreen}\textbf{O-CLIP-\methodshort} & ViT-H-14 & \textbf{.098} & \textbf{.336} & \textbf{.449} & \textbf{.483} & \textbf{.696} & \textbf{.789} \\
\midrule
\cellcolor{ConcreteBlue}EVA-CLIP & EVA02-B-16 & .045 & .151 & .218 & .225 & .412 & .516 \\
\cellcolor{MethodGreen}\textbf{EVA-CLIP-\methodshort} &EVA02-B-16 & \textbf{.052} & \textbf{.175} & \textbf{.249} & \textbf{.258} & \textbf{.463} & \textbf{.564} \\
\hdashline
\cellcolor{ConcreteBlue}EVA-CLIP & EVA02-L-14 & .063 & .217 & .307 & .308 & .521 & .618 \\
\cellcolor{MethodGreen}\textbf{EVA-CLIP-\methodshort}& EVA02-L-14  & \textbf{.072} & \textbf{.252} & \textbf{.345} & \textbf{.352} & \textbf{.574} & \textbf{.676} \\
\hdashline
\cellcolor{ConcreteBlue}EVA-CLIP & EVA02-E-14 & .090 & .312 & .423 & .440 & .667 & .762 \\
\cellcolor{MethodGreen}\textbf{EVA-CLIP-\methodshort} & EVA02-E-14 & \textbf{.096} & \textbf{.334} & \textbf{.451} & \textbf{.471} & \textbf{.689} & \textbf{.787} \\
        \bottomrule
    \end{tabular}
}
\caption{
       Retrieval performance of ~\inlineColorbox{MethodGreen}{\methodshort} on DeepFashion when integrated on different ~\inlineColorbox{ConcreteBlue}{zero-shot} models. %of different families and scales. 
       Shifting towards concrete representations consistently provides a performance boost.\label{tab:exp-scaling}}
\end{table}

\noindent\textbf{Cross-dataset retrieval.}\label{subsec:cross_dataset}
As real-life applications of fashion-domain text-to-image retrieval models should be deployed in the wild, we consider a cross-dataset evaluation where the model is required to generalize beyond its training distribution.
In the bottom block of Tab.~\ref{tab:results-deepfashion}, we report the cross-dataset performance of the different models fine-tuned on FACAD~\citep{yang2020fashion} (\texttt{<model>-ft-facad}), and compare them to \ours with the \AC database built on FACAD. 
\methodshort's gain is consistent over previous state-of-the-art approaches on all retrieval metrics with an average improvement of $8.6\%$ with respect to the second-best zero-shot approach (SigLIP).
Notably, our cross-dataset version achieves almost identical results to the same-dataset scenario (\texttt{\methodshort-df}). On the other hand, we see a consistent performance drop in the fine-tuned backbones when compared to their same-dataset versions. This confirms our hypothesis, highlighting the generalization capability of our representation shift and simultaneously showcasing, once again, the difficulties of popular VLMs when faced with concepts not seen during training. Experiments in the opposite direction (fine-tuning on DeepFashion, testing on FACAD) are reported in the \suppmat

\noindent\textbf{Model scaling.}
With the rapid growth of the VLM field, increasingly powerful models are released daily. We explore how \methodshort can be adapted for integration across different model families and scales.
Tab.~\ref{tab:exp-scaling} reports the retrieval metrics on a same-dataset setting compared to zero-shot models on DeepFashion.
Our results show that \methodshort consistently provides a performance gain across all approaches (with an average of $+4.9\%$ H@1 over all backbones) and model sizes 
(\eg on \texttt{EVA-CLIP}, \methodshort provides an average of $+3.6\%$ H@1 across all model dimensions).
We hypothesize that, due to the concreteness of data used during model pre-training highlighted in Sec.~\ref{sec:preliminary}, larger pre-trained models still under-represent abstract concepts, justifying \methodshort as a necessary approach to deal with the representation shift.

\begin{table}
\centering
\scriptsize
\begin{tabular}{cc cc}
        \toprule
        \textbf{Representation Shift}  & \textbf{Language Rewriting} &  \textbf{R@5} & \textbf{H@1} \\
         \midrule
        \textcolor{red}{\normalsize \ding{55}} & \textcolor{red}{\normalsize \ding{55}} & .228 & .311 \\
        
         \textcolor{red}{\normalsize \ding{55}} & Llama 3.1-8B & .294 & .411 \\
         
         Qwen2-VL & \textcolor{red}{\normalsize \ding{55}} & .246 & .347 \\
         
         \hdashline
         CogVLM2 & Llama 3.1-8B & .302  & .433 \\
         Img-embeddings & Llama 3.1-8B & .266 & .406  \\
         \hdashline
         Qwen2-VL & Llama 3.2-3B & .259 & .365 \\
         Qwen2-VL & Phi3-mini-4K & .295 & .424 \\
        \hline
         \cellcolor{MethodGreen}Qwen2-VL & \cellcolor{MethodGreen}Llama 3.1-8B & \textbf{.303}  & \textbf{.437} \\
        \bottomrule
    \end{tabular}
    \caption{Ablation analysis on the components of \inlineColorbox{MethodGreen}{\methodshort}, with performance on the retrieval task on DeepFashion. 
    In green, the parameters used for our \methodshort. Both the Language Rewriting and Representation Shift steps are needed to get the best performance. Furthermore, \methodshort is robust to different component choices.}\label{tab:ablation}
\end{table}

\noindent\textbf{Ablation analysis.}
We conduct ablation experiments to evaluate the contribution of \methodshort pipeline components. Tab.~\ref{tab:ablation} reports the retrieval performance on DeepFashion.
As can be noted from the first section, both the Representation Shift and the Language Rewriting
provide a benefit to the original model retrieval performance with a +3.7\% and 10\% H@1 gain, respectively. While properly prompted LLMs can provide an effective first estimate of the concrete counterpart of an abstract-oriented description, \methodshort further boosts the performance by explicitly accounting for the representation shift.
In the second section of Tab.~\ref{tab:ablation}, we experiment with alternative sources of concrete information, such as captions from \texttt{CogVLM2}~\citep{hong2024cogvlm2} and the direct use of visual embeddings (\texttt{Img-embeddings}) coming from the VLM image encoder. Results show that on one side \methodshort is robust to captions coming from another state-of-the-art VLM, 
while on the other, moving from textual to visual concrete information leads to performance degradation. We hypothesize the modality gap~\citep{liang2022mind} between visual and text-embedding may dominate the shift. 
Finally, we ablate the language-rewriter choice with different open-source instructed LLMs, including \texttt{Llama-3.1-8B}, \texttt{Llama-3.2-4B} and \texttt{Phi3-mini-4k}.
\methodshort proves robust to the choice of the language rewriter, achieving consistently higher performance over the pre-trained backbones.

%% file: sec/6_conclusions.tex
\section{Conclusions}
\label{sec:conclusions}

Our work advances the exploration of abstract-oriented language in VLMs, focusing on the fashion domain with multimodal datasets. We show that abstract language, though underrepresented in pre-training datasets, is crucial for retrieval tasks. To address this gap, we propose \methodfull{} (\methodshort), a training-free, model-agnostic approach leveraging pre-trained models and abstract-oriented datasets. Our method significantly improves text-to-image retrieval, outperforming fine-tuned VLMs in both same- and cross-dataset settings.

\noindent\textbf{Limitations \& Future work.}
Our study focuses on identifying abstract expressions through adjectives, providing a starting point for advancing research in abstract-oriented language. While we use the fashion domain as a representative case, exploring other fields like arts and movies with similar abstract language would be valuable.

\noindent\textbf{Broader Societal Impacts.}
The analysis and process of fashion data require careful ethical considerations. Necessary measures could include, but are not limited to, anonymization and bias mitigation. 

%% file: sec/acknowledgements.tex
\section{Acknowledgment}
This study was supported by LoCa AI, funded by Fondazione CariVerona (Bando Ricerca e Sviluppo 2022/23), PNRR FAIR - Future AI Research (PE00000013) and Italiadomani (PNRR, M4C2, Investimento 3.3), funded by NextGeneration EU.
We acknowledge the CINECA award under the
ISCRA initiative, for the availability of high-performance computing resources and support.
We acknowledge EuroHPC Joint Undertaking for awarding us access to MareNostrum5 as BSC, Spain.
Finally, we acknowledge HUMATICS, a SYS-DAT Group company, for their valuable contribution to this research.

%% file: supp/supp-preamble.tex
\clearpage
\clearpage
\setcounter{page}{1}
\setcounter{figure}{0}
\setcounter{table}{0}
\setcounter{equation}{0}
\setcounter{section}{8}
\renewcommand{\thesection}{\AlphAlph{\value{section}-8}}

\maketitlesupplementary

%% file: supp/intro.tex
\section{Overview}
In this supplementary, we provide additional details on the implementation of the presented approach and additional experimental analysis on \ours. In Section~\ref{supp:additional-details} we present further details on the dataset discussed in our work, the attribute extraction pipeline for the preliminary analysis, and extensive details on \ours. Section~\ref{supp:additional-analysis} presents additional analysis on the latent space of the Vision Language Model (VLM) and the cross-dataset evaluation on FACAD. Finally, in Section~\ref{supp:qualitatives}, we show additional interesting qualitative results of the retrieval on DeepFashion using~\ours. 

%% file: supp/further-details.tex
\section{Additional details}
\label{supp:additional-details}
In this section we provide additional details on the dataset used for the presented analysis (Sec.~\ref{supp:additiona-details-dataset}) and report implementation details, especially focusing on the extraction pipeline leveraged for fashion language analysis (Sec.~\ref{supp:extraction-pipeline}) and prompting information (Sec.~\ref{supp:captioning-rewriting-details}).
\subsection{Fashion datasets vs LAION 400M}
\label{supp:additiona-details-dataset}

\begin{figure}[t!]
    \centering
    \includegraphics[width=0.94\linewidth]{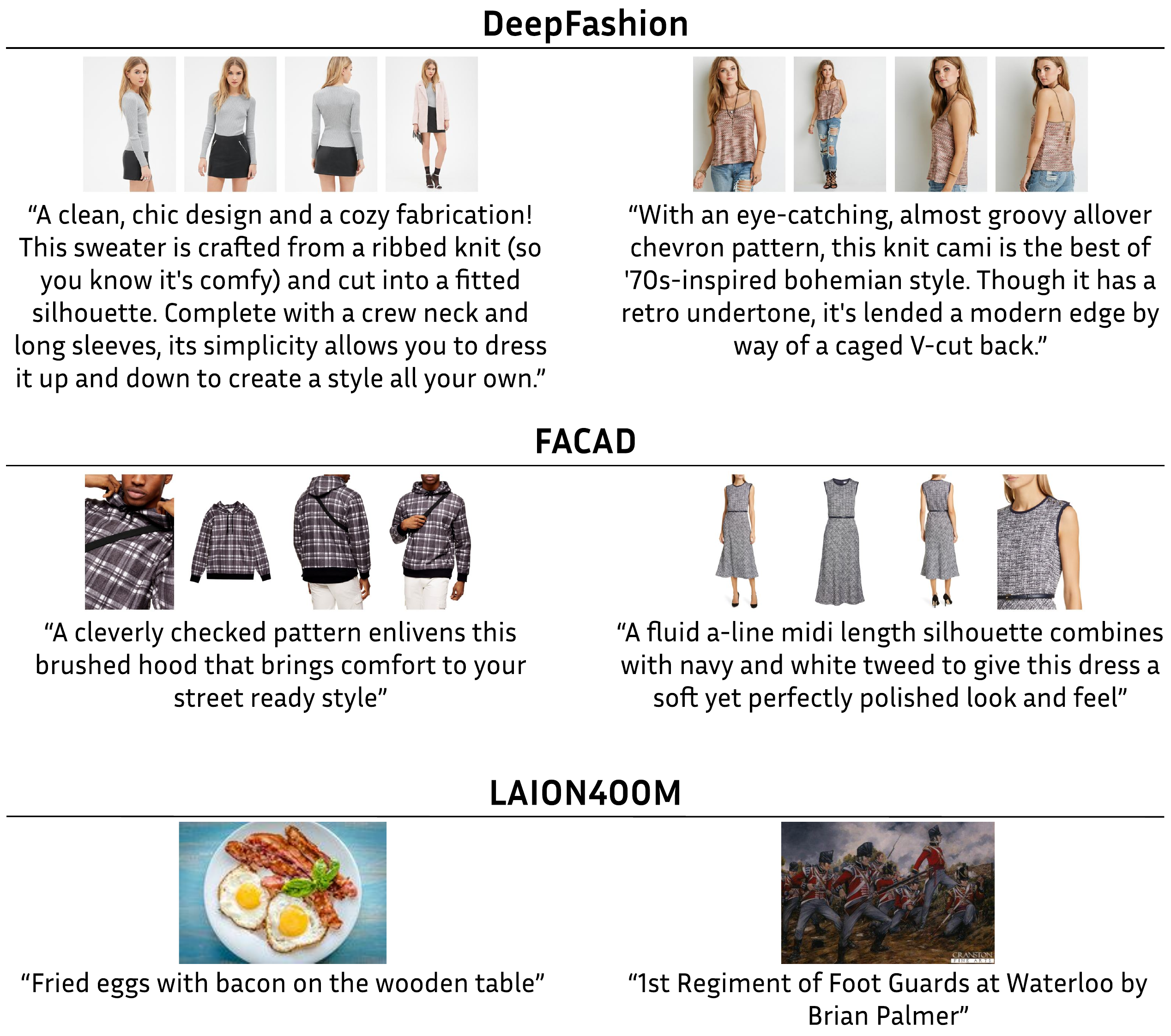}
    \caption{Samples taken from the datasets considered for analysis in our main paper. Compared to the abstract-oriented fashion datasets, LAION 400M mainly provides short and concrete details about the image content.}
    \label{fig:dataset_samples}
\end{figure}

\begin{figure}[t!]
    \centering
    \includegraphics[width=\linewidth]{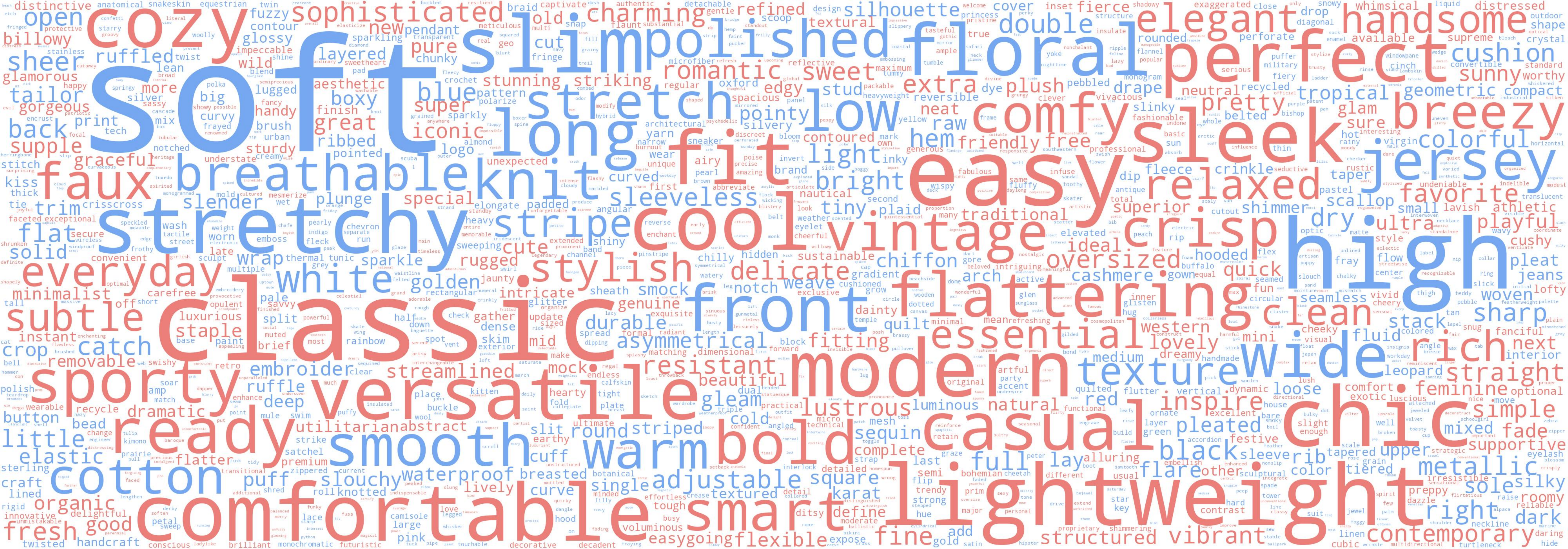}
    \caption{Wordcloud of \inlineColorbox{ConcreteBlue}{concrete} (blue) and \inlineColorbox{AbstractRed}{abstract} (red) adjectives in FACAD descriptions. The larger font indicates a higher frequency. Abstract adjectives make up a large portion of the entire dataset, being as frequent as concrete ones (see \cref{tab:adj_freq} of the main paper).}
    \label{fig:facad_wordcloud}
\end{figure}

\begin{figure}[t!]
    \centering
    \includegraphics[width=\linewidth]{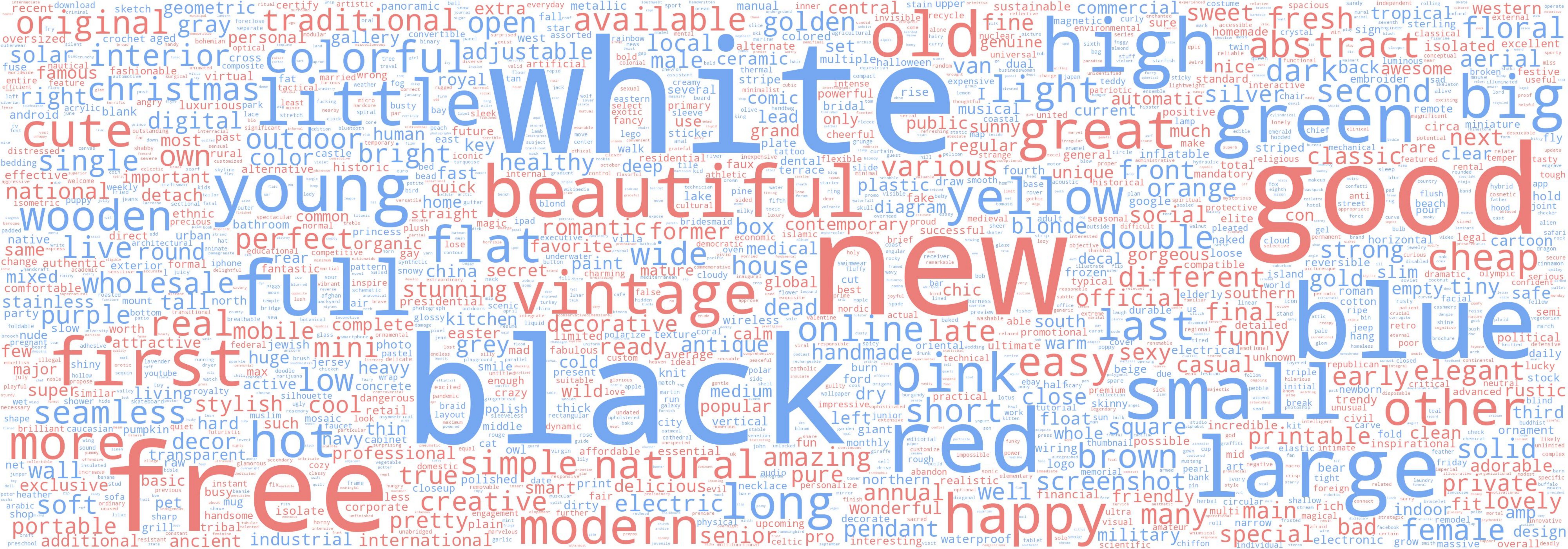}
    \caption{Wordcloud of \inlineColorbox{ConcreteBlue}{concrete} (blue) and \inlineColorbox{AbstractRed}{abstract} (red) adjectives in LAION descriptions. The larger font indicates a higher frequency. It's clear how concrete adjectives are more frequent than abstract ones (see \cref{tab:adj_freq} of the main paper).}
    \label{fig:laion_wordcloud}
\end{figure}

 In \cref{fig:dataset_samples} we show some data samples taken from the fashion-related DeepFashion and FACAD, as well as two samples randomly picked from LAION 400M. It is clear how, apart from the image content, the descriptions differ greatly: both DeepFashion and FACAD provide longer descriptions with more abstract-oriented properties, such as ``chic'' and ``street ready'', to describe both appearance and feeling precisely; on the other hand, LAION 400M focuses the descriptions on the visually grounded aspects to briefly describe the content. This difference is further highlighted in the adjective wordclouds of FACAD~\cref{fig:facad_wordcloud} and LAION~\cref{fig:laion_wordcloud}. While FACAD has a balanced distribution between concrete and abstract adjectives, LAION 400M exhibits a noticeable bias towards concrete adjectives.
 The word cloud is predominantly filled with concrete descriptors with a strong emphasis on color-related terms like ``white'' and ``black''. 
 For exact statistical numbers, we refer the reader to \cref{tab:adj_freq} in the main paper.

\subsection{Analysis extraction pipeline}
\label{supp:extraction-pipeline}
Our attribute extraction and classification pipeline uses the Python-based spaCy library to tokenize, tag, and extract adjectives and attributes from the item descriptions.
The pipeline follows 4 main phases:
\textit{i) preparation phase} where we instantiate the NLP model and tokenizer;
    \textit{ii) tagging phase} where the NLP model performs POS tagging;
    \textit{iii) extraction phase} where we extract the adjectives/attributes from the descriptions;
    \textit{iv) classification phase} where we classify the adjectives/attributes into abstract and concrete following a lexicon.
We briefly discuss the pipeline in the following paragraphs and will release the code upon acceptance.

\noindent\textbf{i) Preparation phase.}  
We first instantiate a language processor using the pre-trained spaCy \texttt{en\_core\_wb\_sm} model and then modify its tokenizer,  keeping hyphenated words as a single token (\eg \texttt{mind-blowing} will be considered a single word, instead of \texttt{[mind, blowing]}). This choice is guided by our observation that hyphenated words are common in the fashion language to convey particular details, \eg \texttt{v-neckline}, \texttt{must-have}, \texttt{happy-mood}, etc.

\noindent\textbf{ii) Tagging phase.}
With the spaCy\citep{spacy} model configured, we then process each description by \textit{tokenizing} and \textit{tagging} them. 
We use the tokenizer introduced in the previous step to parse the description and separate the words into tokens \ie atomic words in the sentence,
then we conduct POS tagging using the spaCy model. This step takes the list of tokens and analyzes their syntactic relationship, building a Part-of-Sentence (POS) tree. In this tree, each token is associated with a tag (\eg adjective, noun, verb, etc.) and connected to other tokens through syntactic relationships, \eg \texttt{green jacket} will tag \texttt{green} as an adjective, and connect it to \texttt{jacket} through an \texttt{adjective modifier} relationship. We will refer to this tree as our \texttt{tagged description}, for simplicity.

\noindent\textbf{iii) Extraction phase.}
With the description successfully tagged, we can then extract the attributes. In particular, when analyzing the adjectives only (\ie dataset statistics in \cref{sec:preliminary} of the main paper), we keep all the words that were tagged as ``adjectives'' or that had an ``adjective modifier'' relationship in the tagged description. 
Finally, to avoid word repetitions, we lemmatize the words using the spaCy model built-in lemmatizer.

Similarly, when analyzing attributes (which we define as adjective+noun couples), we follow the same procedure to first locate all adjective tokens. To extract the attribute, 
we then check whether each of these tokens constitutes a valid attribute with its relationship head (the word it refers to or modifies).
To do so, we apply the following rules: 
\begin{enumerate}
    \item If the head is either an ``adjective'', or was tagged as the ``subject'' of the sentence, lemmatize it and create the attribute (token, head).
    \item If the token was a ``compound word'' (\ie a concept expressed through multiple words, such as ``knee length''), extract all the (token, compound) couples as attributes.
\end{enumerate}

\noindent\textbf{iv) Classification phase.}
Finally, we classify the attributes following the methodology and lexicon presented in~\cite{brysbaert2014concreteness}. 
It is worth noting that some adjectives were not present in the lexicon, and we found many words' concreteness was context-dependent. Words like \texttt{cool} and \texttt{sharp} were rightfully considered concrete in the lexicon since, in a common context, they refer to the ``sensation of cold'' and the ``ability to cut'' respectively, which are properly assessable with senses. In the fashion domain, however, these words change in meaning and rather describe the style of the clothing, which we define as abstract. We manually assigned concrete/abstract classes to these words following suggestions from domain experts.

\noindent\textbf{Examples.}
We provide some examples of the attribute extraction results. In blue we highlight the \inlineColorbox{ConcreteBlue}{concrete} attributes extracted, while in red the \inlineColorbox{AbstractRed}{abstract} ones.
\begin{myquote}
    A duo of \inlineColorbox{ConcreteBlue}{dark stripes} elongate a midweight \inlineColorbox{ConcreteBlue}{cotton polo}, styled in a \inlineColorbox{AbstractRed}{comfortable fit}, with \inlineColorbox{ConcreteBlue}{long sleeves}, and a \inlineColorbox{AbstractRed}{sporty zipper} in the placket.
\end{myquote}

\begin{myquote}
    Stay \inlineColorbox{AbstractRed}{comfy} and \inlineColorbox{AbstractRed}{chic} while showing off your bump in this \inlineColorbox{ConcreteBlue}{scalloped neck sheath}, featuring \inlineColorbox{AbstractRed}{flattering seam}, and a \inlineColorbox{ConcreteBlue}{knee length skirt}.
\end{myquote}

\subsection{Prompt details}
\label{supp:captioning-rewriting-details}
\noindent\textbf{Captioning.} For captioning purposes, we utilize recent state-of-the-art instructed VLMs, referred as captioning models, allowing for a detailed description of the fashion item of interest. Specifically, we experiment with Int4-quantized versions of Qwen2-VL~\citep{wang2024qwen2} (\texttt{Qwen2-VL-7B-Instruct-GPTQ}) and CogVLM2~\citep{hong2024cogvlm2} (\texttt{cogvlm2-llama3-chat-19B}) that are queried to focus on item class using the vision prompt $p_v$:
\begin{myquote}
    $p_v$: Briefly describe in detail the $\langle\texttt{class}\rangle$ only in less than 10 tokens, use visually grounded properties like shape, pattern and garment details.
\end{myquote}
where $\langle\texttt{class}\rangle$ denotes the class of the fashion item of interest. The caption generation is limited to 77 tokens.

\noindent\textbf{Language rewriting.} Language rewriting relies on the use of a LLM for a first rephrased version of the provided abstract query in more concrete terms. For the main experimentation, we use an instructed version of Llama-3~\citep{dubey2024llama} (\texttt{Llama-3.1-8B-Instruct}) loaded for inference in half precision. The model is guided to focus on concrete aspects of the original descriptions, discarding information that downstream VLMs cannot leverage for matching. In practice, we use the prompt:
\begin{myquote} 
$p_r$: Briefly describe the fashion item in less than 10 tokens in a natural language sentence in a discriminative way. Avoid specialized language and listing properties. Focus on shape, patterns, color (if indicated) and garment details. Remove information on how to pair and complement. Strictly remove suggestions on occasions when to wear it.
\end{myquote}
We further add \emph{four} In-Context samples~\citep{dong2022survey} to provide demonstrations of the desired rewriting outcomes:
\begin{myquote}
This sheer Georgette top features a high collar and shirred shoulders. It features long sleeves and buttoned cuffs $=\rangle$ A sheer, high-collared Georgette top with shirred shoulders, long sleeves, and buttoned cuffs.
\end{myquote}
\begin{myquote}
Represent your love for Long Beach hip-hop with this old school graphic of Snoop Dogg on a muscle tee. To balance some softness into this edgy piece, tuck it into a skater skirt and finish it off with flatforms $=\rangle$ A graphic muscle tee featuring Snoop Dogg.
\end{myquote}
\begin{myquote}
Busy mornings and stacked social calls are no match for this throw-on-and-go tunic! Its boxy silhouette and ultra-soft knit fabrication will have you taking on the day in breezy comfort, while a scoop back lends it some unexpected sartorial side-eye. Plus, you can just as easily throw on this short-sleeved number over leggings to effortlessly transition from day to night. $=\rangle$ A casual, boxy tunic with a scoop back, made from soft knit fabric.
\end{myquote}
\begin{myquote}
Cut from a crisp cotton woven splashed with a richly ornate paisley print, this short-sleeved shirt is guaranteed to be a standout in your collection. Its boldness is pared down by its sleek and refined structure (a slim fit, classic collar, and buttoned front). Just because it's a standout doesn't mean it isn't versatile - wear it with everything from chinos to joggers for a sharp dose of style. $=\rangle$ A crisp cotton shirt with a paisley print, slim fit, classic collar, and buttoned front.
\end{myquote}
The provided rewritten descriptions are randomly drawn from the DeepFashion evaluation set and hence removed from the final performance testing. These sample descriptions are the only captions manually re-written with human involvement.
A similar prompt is used with other LLMs when required to ablate on the language rewriting choice, \eg the smaller models Llama-3.2~\citep{dubey2024llama}  (\texttt{Llama-3.2-3B-Instruct}) and Phi-3~\citep{abdin2024phi} (\texttt{Phi-3-mini-4k-instruct}) that are loaded in full precision.
Due to the complexity of brief and ungrammatical FACAD descriptions, for FACAD cross-dataset setting we refine the prompt to maintain a similar style and description length:
\begin{myquote} 
$p_r$: Briefly rewrite in a discriminative way to focus on shape, patterns, color (if indicated) and garment details, preserve all information. Use a single sentence with description length and writing style similar to original descriptions. Remove information on how to pair and complement. Strictly remove suggestions on occasions when to wear it.
\end{myquote}
and present results relying on a Int4-quantized \texttt{Phi-3-medium-4k-instruct}, based on empirical observations.

Tab.~\ref{tab:llm-inout} and Tab.~\ref{tab:llm-facad-inout} reports qualitatives on language rewriting for DeepFashion and FACAD, respectively. As can be noted, while for DeepFashion language rewriting allows to focus on concrete-information about the fashion item of interest, FACAD rewriting mostly focus on correcting the description grammar-wise.

\begin{table*}[h!]
\centering
\begin{tabular}{l p{13.5cm}}
    \toprule
    \textbf{Original query ($q^A$)} & This floral print maxi dress features a front slit and self-tie crossback straps. Team this piece with flat sandals and sunnies for a warm weather look that all ease.\\
    \textbf{Language rewritten ($q'$)} & A floral print maxi dress with a front slit and self-tie crossback straps.\\
    \midrule
    \textbf{Original query ($q^A$)} & Crafted from a stretchy ribbed knit with a racerback design and the text  ``Where There's a Will, There's a Way'', this sporty tank is a shortcut to street style that makes a statement (literally). Plus, its slightly cropped cut means it's the cute partner in crime your high-waisted bottoms have been missing.\\
    \textbf{Language rewritten ($q'$)} & A sporty tank top with a racerback design and a printed message, made from stretchy ribbed knit, featuring a slightly cropped cut.\\
    \midrule
    \textbf{Original query ($q^A$)} & There are certain pieces that will always bring a boho style aesthetic to mind, and this boxy top is one of them. It's crafted from an open floral crochet with a scalloped hem and short sleeves. We're showcasing the circle crochet trim along the round neckline with a turquoise necklace to really knock it out of the park. \\
    \textbf{Language rewritten ($q'$)} & A boxy top with an open floral crochet pattern, scalloped hem, and short sleeves, featuring a circle crochet trim along the round neckline.\\
    \midrule
    \textbf{Original query ($q^A$)} & The easiest way to take-on a rushed morning starts with this heathered top! Cut from an ultra-lightweight fabric into a slouchy fit, make this long-sleeved piece will be your next go-to. With it's classic design, this layer will seamlessly work into any outfit for a relaxed aesthetic. \\
    \textbf{Language rewritten ($q'$)} & A slouchy, long-sleeved top made from ultra-lightweight fabric.\\
    \midrule
    \textbf{Original query ($q^A$)} & Cookie-cutter so isn't your style. This muscle tee gets you with its ``I'm Original'' graphic, raw-cut hem, and cutout back. Rock a lacy bandeau underneath and put your unique spin on the look.\\
    \textbf{Language rewritten ($q'$)} & A muscle tee featuring an ``I'm Original'' graphic, a raw-cut hem, and a cutout back.\\
    \bottomrule
\end{tabular}
\caption{Examples of input abstract-oriented descriptions and their corresponding LLM rewritten descriptions, using \texttt{Llama-3.1-8B-Instruct} on DeepFashion. All inferences rely on the same $p_r$ prompt, see Sec.~\ref{supp:captioning-rewriting-details}.\label{tab:llm-inout}}
\end{table*}

\begin{table*}[h!]
\centering
\begin{tabular}{l p{13.5cm}}
    \toprule
    \textbf{Original query ($q^A$)} & chewed hem amp up the vintage vibe of these high waisted stretch denim jeans featuring 3d whisker minor abrasion and hand sanding through the thigh\\
    \textbf{Language rewritten ($q'$)} & high-waisted stretch denim jeans with a chewed hem, 3D whisker abrasion, and hand-sanded thighs for a vintage look.\\
    \midrule
    \textbf{Original query ($q^A$)} & an ortholite footbed and cushion soft padding ensure all day comfort in a versatile chukka boot in a sharp silhouette\\
    \textbf{Language rewritten ($q'$)} & a versatile chukka boot with a sharp silhouette, featuring an ortholiste footbed and cushion soft padding for all-day comfort.\\
    \midrule
    \textbf{Original query ($q^A$)} & ripped hem and heavily sanded highlight bring instant old favorite status to a vintage button up that s easy to wear layered or on it own\\
    \textbf{Language rewritten ($q'$)} & a vintage button-up shirt with a ripped hem and heavily sanded highlight, versatile for layering or wearing alone.\\
    \midrule
    \textbf{Original query ($q^A$)} & a faux wrap top cut in a cool linen blend feature a wide sash at the waist that creates a pretty peplum effect\\
    \textbf{Language rewritten ($q'$)} & a faux wrap top in a cool linen blend with a wide sash at the waist for a peplum effect\\
    \midrule
    \textbf{Original query ($q^A$)} & a slit at the front brings easy movement to this stretch denim skirt finished with a softly fraying hem\\
    \textbf{Language rewritten ($q'$)} & a stretch denim skirt with a front slit and a softly fraying hem\\
    \bottomrule
\end{tabular}
\caption{Examples of input abstract-oriented descriptions and their corresponding LLM rewritten descriptions, using \texttt{Phi-3-medium-4k-instruct} on FACAD. All inferences rely on the same $p_r$ prompt, see Sec.~\ref{supp:captioning-rewriting-details}. \label{tab:llm-facad-inout}}
\end{table*}

%% file: supp/additional-exp.tex
\section{Additional Analysis}
\label{supp:additional-analysis}
In this section we present additional analysis on our presented approach~\ours. 
Sec~\ref{supp:latent-space} explores the representation shift that occurs, while Sec.~\ref{supp:cross-dataset-facad} presents a quantitative evaluation of the FACAD cross-dataset analysis.
\begin{figure}[t!h!]
    \centering
    \includegraphics[width=0.9\linewidth]{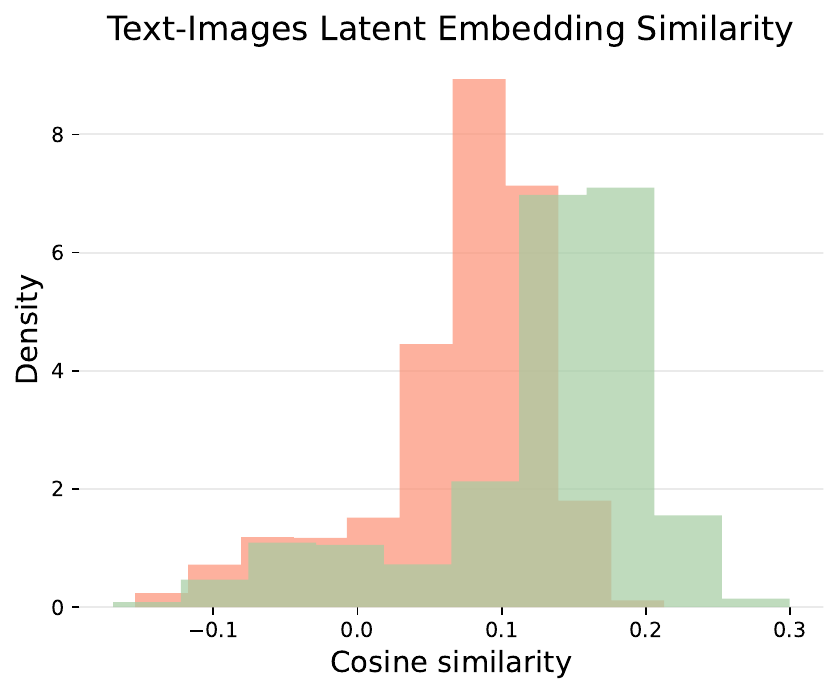}
    \caption{Density distribution of the cosine similarity with respect to the ground truth image embeddings of original\inlineColorbox{AbstractRed}{abstract} queries embeddings and \inlineColorbox{MethodGreen}{\methodshort} ones.}
    \label{fig:latent-sim-distr}
\end{figure}
\subsection{VLM latent space}
\label{supp:latent-space}
\begin{figure*}[h!]
\centering
\includegraphics[width=0.9\linewidth]{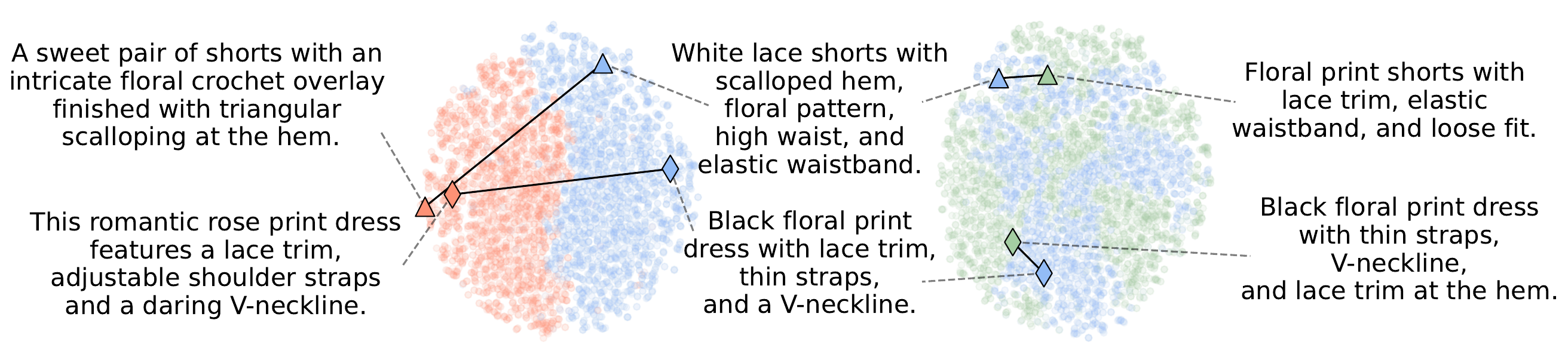}
\caption{t-SNE of VLM latent space for textual embeddings in the evaluation set of DeepFashion. Connected markers denote different descriptions associated to the same data sample. \textbf{Left:} \inlineColorbox{AbstractRed}{Abstract} queries embeddings are separated from \inlineColorbox{ConcreteBlue}{concrete} ones, as those obtained from captioning models, \texttt{Qwen2-VL} in this case. \textbf{Right:} \inlineColorbox{MethodGreen}{\methodshort} allows to reduce the separation in the latent space.}
 \label{fig:latent-tsne}
\end{figure*}
We investigate the latent transformations induced by \ours on representations of abstract textual descriptions. To assess this, we compute the cosine similarity between embeddings of original abstract descriptions and their corresponding ground truth visual representations within a shared latent space. These similarities are then compared to those obtained after applying the \ours transformation to the original abstract descriptions, see Fig.~\ref{fig:latent-sim-distr}. The results demonstrate a clear improvement in alignment, as evidenced by a shift in the similarity distribution toward higher values relative to the baseline. This suggests that \ours enhances the semantic correspondence between text and images. 

To further examine the structure of the latent space, we employ t-SNE~\citep{van2008visualizing} to visualize the embeddings of textual descriptions of the retrieval VLM. Figure~\ref{fig:latent-tsne}
 contrasts the original and shifted embeddings with those associated with concrete representations derived from a captioning model. The visualization reveals that, while embeddings from abstract descriptions are initially separated from their concrete counterparts, the \ours-transformed embeddings exhibit improved mixing. This blending demonstrates the establishment of a shared representation space where textual representations of abstract and concrete descriptions overlap.
 To show that \ours{} effectively shifts in the language space, we report the textual descriptions corresponding to marked embeddings, where we can observe a clear coherence in concrete terms.
 The results indicate that \ours effectively bridges the gap between abstract and concrete descriptions of the same fashion data, leading to unified multimodal understanding.

\subsection{Cross-dataset evaluation with FACAD}
\label{supp:cross-dataset-facad}
\begin{table}[h!]
\footnotesize
    \centering
    \resizebox{\linewidth}{!}{    
    \begin{tabular}{l ccc ccc ccc}
        \toprule
        \textbf{Model} & \textbf{R@1} & \textbf{R@5} &\textbf{R@10} & \textbf{H@1} & \textbf{H@5} &\textbf{H@10} \\ 
        \midrule
        \multicolumn{7}{c}{\cellcolor{Gray!16}\textbf{Zero-Shot}}\\
      \cellcolor{ConcreteBlue}SigLIP          & 0.085	& 0.291	& 0.404	& 0.441	& 0.678	& 0.766 \\
      \cellcolor{ConcreteBlue}CLIP           & 0.022 & 0.073 & 0.109 & 0.107 & 0.247 & 0.333 \\
      \cellcolor{ConcreteBlue}O-CLIP         & 0.064 & 0.207 & 0.292 & 0.332 & 0.576 & 0.671 \\
      \cellcolor{ConcreteBlue}EVA-CLIP       & 0.054 & 0.176 & 0.254 & 0.277 & 0.494 & 0.597 \\
      \cellcolor{ConcreteBlue}F-CLIP         & 0.063 & 0.206 & 0.288 & 0.322 & 0.554 & 0.654 \\
      \cellcolor{ConcreteBlue}OF-CLIP        & 0.058 & 0.183 & 0.259 & 0.295 & 0.529 & 0.641 \\
      
    \midrule
    \multicolumn{7}{c}{\cellcolor{Gray!16}\textbf{Cross-Dataset (DeepFashion $\rightarrow$ FACAD)}}\\
\cellcolor{AbstractRed}SigLIP-ft-df      & \textbf{0.093}	& \textbf{0.322}	& \textbf{0.438}	& \textbf{0.482}	& \textbf{0.708}	& \textbf{0.786} \\   
        \cellcolor{AbstractRed}CLIP-ft-df        & 0.023 & 0.076 & 0.116 & 0.115 & 0.257 & 0.349 \\
       \cellcolor{AbstractRed}O-CLIP-ft-df      & 0.066 & 0.213 & 0.299 & 0.341 & 0.584 & 0.680 \\
       \cellcolor{AbstractRed}EVA-CLIP-ft-df    & 0.051 & 0.171 & 0.244 & 0.265 & 0.477 & 0.576 \\
       \cellcolor{AbstractRed}F-CLIP-ft-df & 0.063 & 0.207 & 0.289 & 0.328 & 0.559 & 0.649 \\ 
       \cellcolor{AbstractRed}OF-CLIP-ft-df & 0.062 & 0.201 & 0.282 & 0.316 & 0.570 & 0.666 \\
        \cellcolor{MethodGreen}\textbf{\methodshort-df (Ours)}         & 0.082 & 0.284 & 0.388 & 0.426 & 0.655 & 0.743 \\
      \cellcolor{MethodGreen}\textbf{\methodshort-df w/o LR (Ours)}      & \underline{0.089} & \underline{0.303} & \underline{0.413} & \underline{0.460}	& \underline{0.694}	& \underline{0.776} \\
\bottomrule
    \end{tabular}
            }
    \caption{
       Results on FACAD. In \textbf{bold} the best results, while \underline{underlined} are the second best. \inlineColorbox{MethodGreen}{\methodshort} proves to be second best method \textit{w.r.t.} ~\inlineColorbox{ConcreteBlue}{zero-shot} and \inlineColorbox{AbstractRed}{fine-tuned} models in the cross-dataset setting where DeepFashion is used for fine-tuning or database construction.} \label{tab:results-facad}
\end{table}
We conducted additional analysis on a cross-dataset setting where models are trained or utilize data from DeepFashion and are evaluated on a 5K subsample of FACAD dataset. Tab.~\ref{tab:results-facad} shows the retrieval performance of zero-shot models, models fine-tuned on DeepFashion and two different variants of the proposed \ours. On one side, \texttt{\ours-df} considers both the language rewriting and representation shift components of the strategy. On the other, \texttt{\ours-df w/o LR} evaluates the representation shift without language rewriting. 
The results indicate that the shift characterization effectively bridges the representation of abstract-oriented descriptions, achieving an average improvement of +1.2\% compared to the best-performing zero-shot model.
\begin{figure*}[h!t!]
    \centering
    \includegraphics[width=0.90\linewidth]{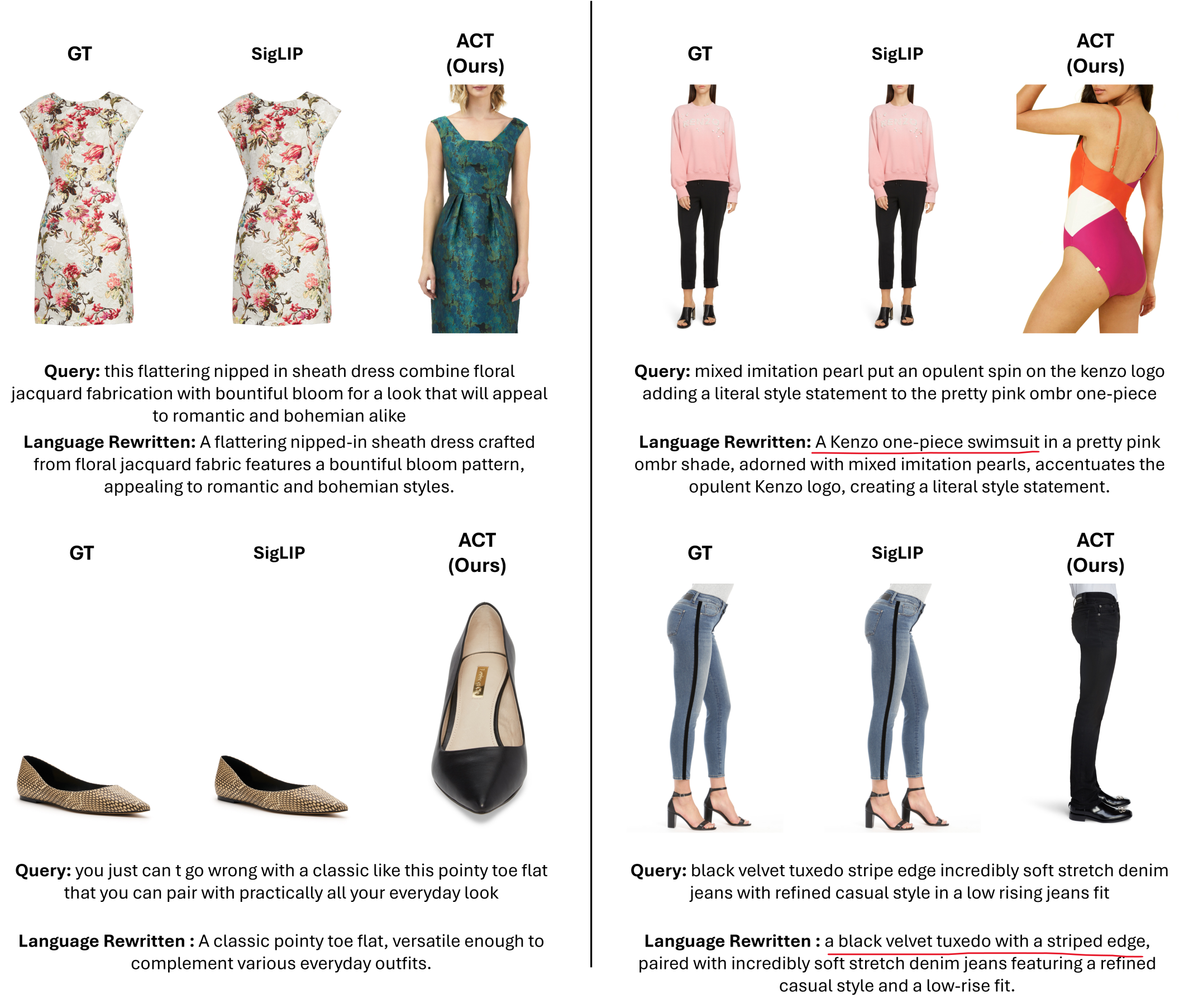}
    \caption{Some failure cases of \ours on FACAD. On the left, we report failures due to the generic FACAD descriptions: despite \ours retrieving fitting candidate images, the retrieval is considered a failure because the exact item was not presented. On the right, we report failures due to language rewriting hallucinations, highlighting the hallucinated elements.}
    \label{fig:llm-qualitative-investigation}
\end{figure*}
The strategy consistently achieves second-best results falling behind the fine-tuned SigLIP model. However, challenges arise when applying large language models (LLMs) for language rewriting, as the performance deteriorate \textit{w.r.t} the zero-shot model. 
A further qualitative investigation, shown in Fig.~\ref{fig:llm-qualitative-investigation}, reveals that the concise and grammatically intricate nature of the original FACAD descriptions (21 words on average per description in FACAD, compared to 53 words on average in DeepFashion) often leads LLMs to hallucinate or lack critical information, reducing their effectiveness in bridging the shift between concrete-oriented and abstract-oriented descriptions. 
This limitation not only impacts performance but also opens promising research directions in the natural language processing fields, particularly in designing LLMs better equipped in understanding and processing attribute-rich, concise text.

%% file: supp/qualitatives.tex
\section{Qualitatives results}
\label{supp:qualitatives}
\begin{figure*}[h!t!]
    \centering
\includegraphics[width=0.80\linewidth]{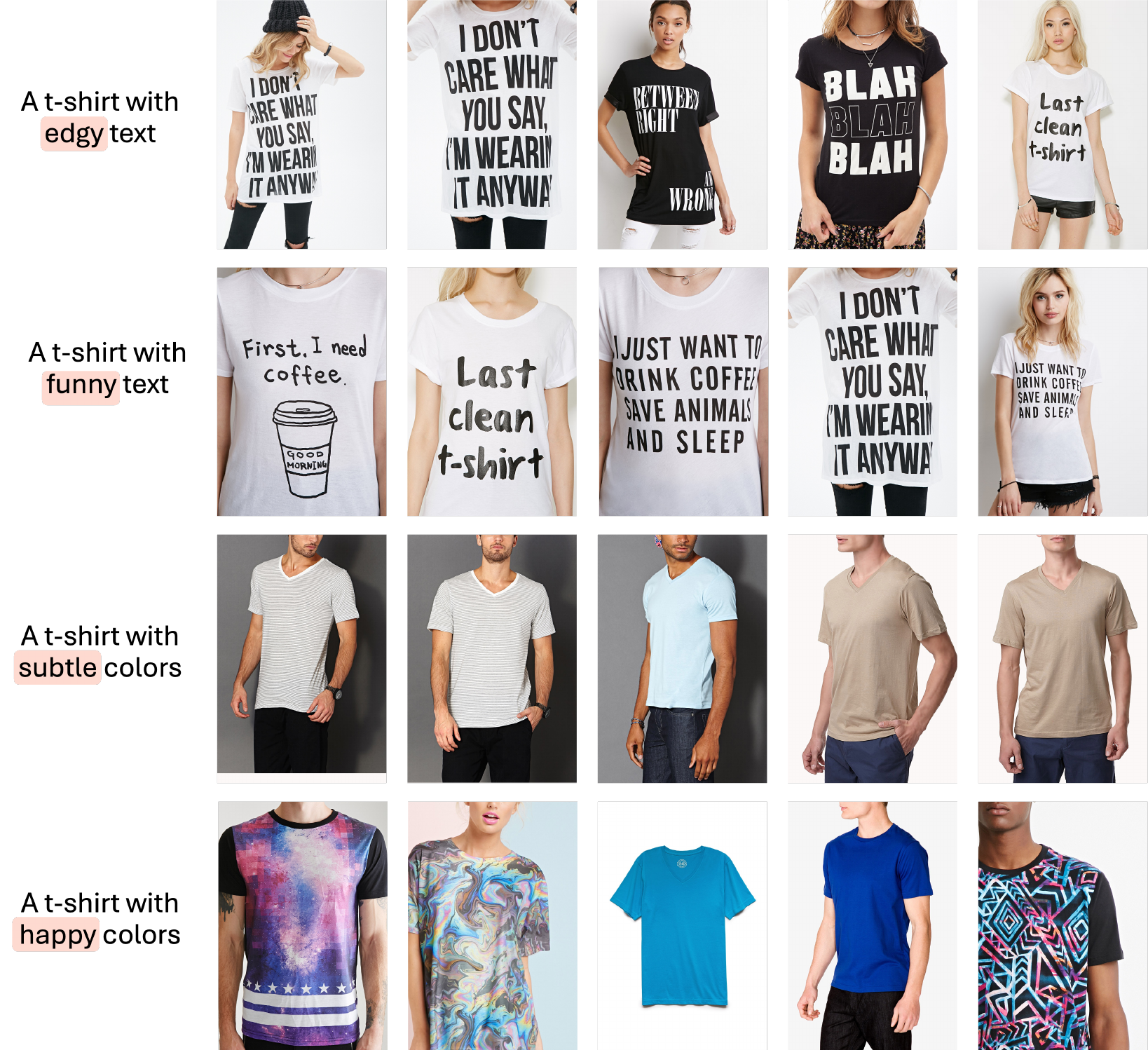}
\caption{Qualitative examples of retrieval using our \methodshort, on the test split of DeepFashion, using abstract attributes. In this example, we provide specialization over different types of t-shirts. As demonstrated, \ours allows us to specialize our retrieval by changing the abstract adjectives used (\eg “subtle colors” vs. “happy colors”).}
\label{fig:qualitatives01}
\end{figure*}
\begin{figure*}
    \centering
    \includegraphics[width=0.85\linewidth]{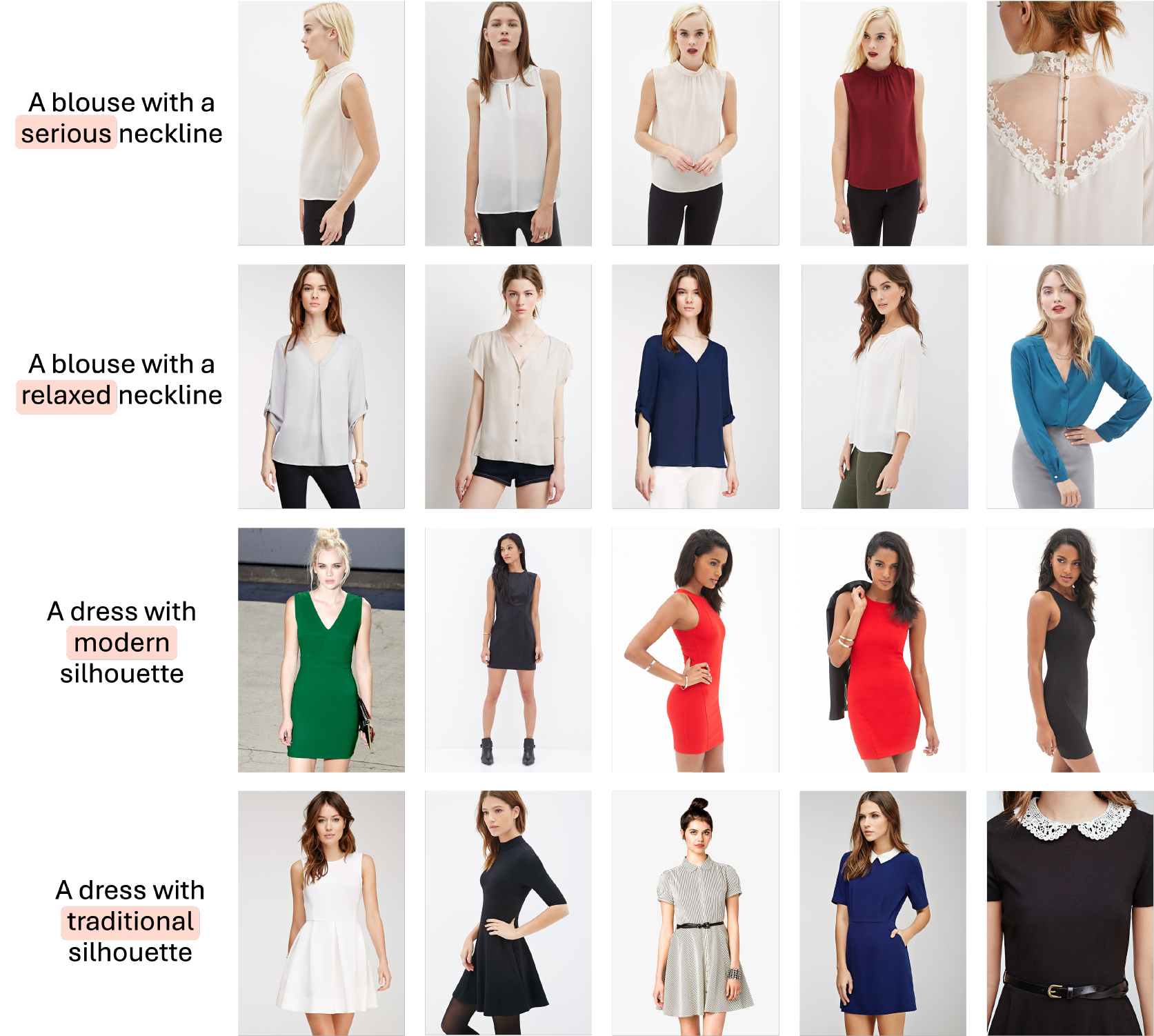}
    \caption{Further qualitative examples of retrieval using our \methodshort, on the test split of DeepFashion, using abstract attributes. In the top two rows, \ours differentiates between necklines, presenting candidate images with tight or high collars for ``serious'' styles, and large, soft v-necklines for ``relaxed'' ones. In the bottom two rows, \ours retrieves sleeveless, short and fitted dresses for a ``modern silhouette'', and mostly sleeved dresses with typically longer, airy skirts for a ``traditional silhouette''}
    \label{fig:qualitatives02}
\end{figure*}
\begin{figure*}
    \centering
    \includegraphics[width=0.85\linewidth]{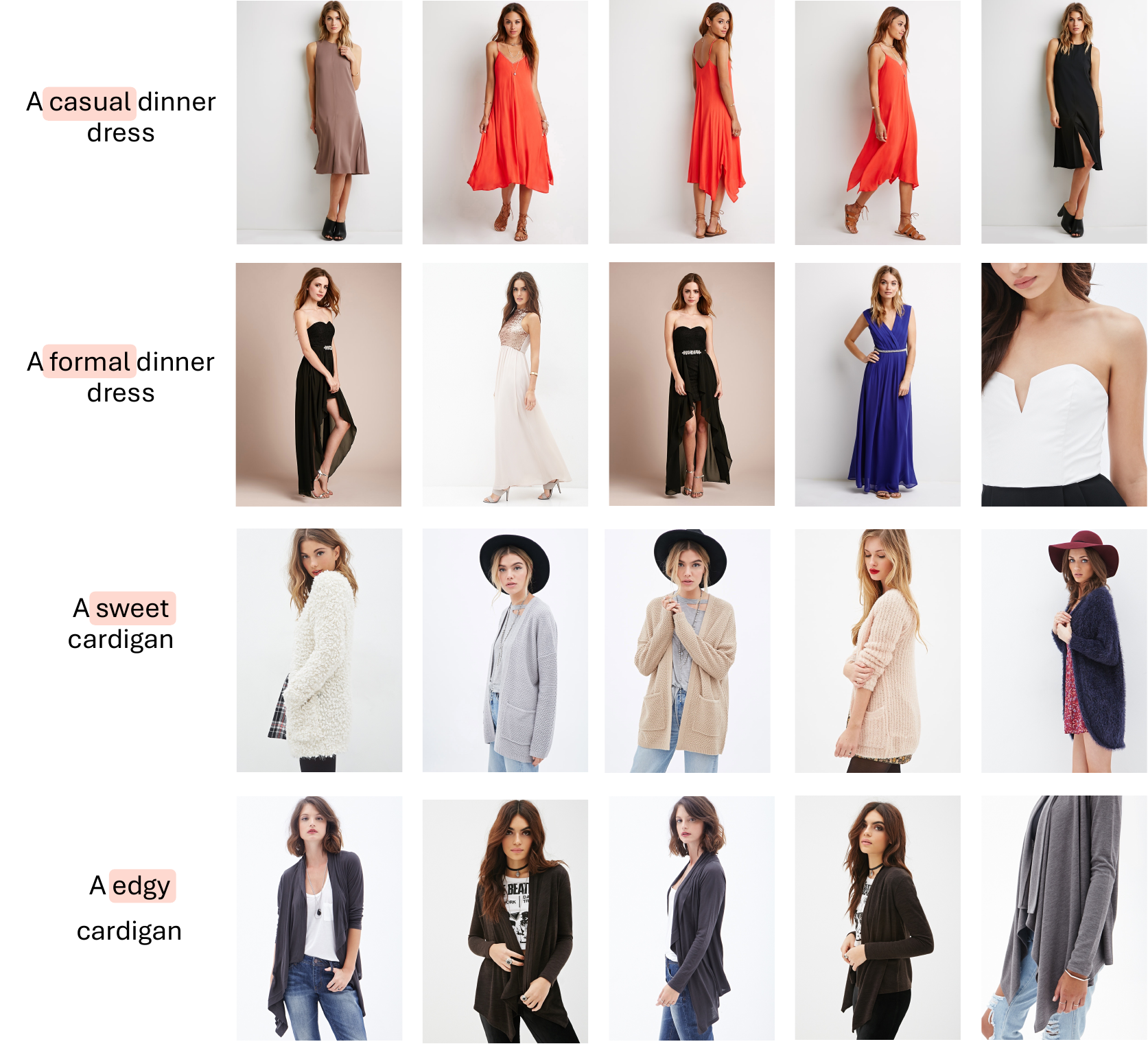}
    \caption{Further retrieval results from our \methodshort on the test split of DeepFashion showcase that \ours can also interpret abstract global details such as styles well. In the top two rows, \ours distinguishes between ``casual'' and ``formal'' dinner dresses, with the former being mono-colored and minimalist, and the latter featuring mostly maxi-length and more sophisticated design. In the bottom two rows, a similar distinction is made between ``sweet'' and ``edgy'' cardigans, with the former having a looser fit and the latter being darker with tighter sleeves.}
    \label{fig:qualitatives03}
\end{figure*}

In this section, we provide some qualitative results of text-to-image retrieval using \ours. 
The queries, specifically designed to qualitatively evaluate the retrieval performance of \ours, are manually crafted to contain a single abstract attribute, and as such, ground truth annotations are not available.
In \cref{fig:qualitatives01}, we illustrate some query specializations across different types of t-shirts. \ours can properly leverage the abstract attributes available (\eg ``subtle colors'' \textit{vs.} ``happy colors'') to present suitable candidates during retrieval. 
In \cref{fig:qualitatives02}, ``serious neckline'' results in blouses with a tight or high collar; meanwhile, when queried with a ``relaxed neckline'', \ours presents candidate images with large and soft v-necklines. Similarly, when asked to retrieve a dress with a ``modern silhouette'', \ours retrieves sleeveless short dresses with a tight fit. A ``traditional silhouette'', on the other hand, leads to mostly sleeved dresses with typically longer, airy skirts.
In \cref{fig:qualitatives03}, we further demonstrate that \ours can interpret styles precisely: when queried for a ``casual dinner dress'', \ours presents mono-colored one-piece dresses that is uncluttered with little-to-no accessories; a ``formal dinner dress'', while also featuring mono-colored, results in mostly maxi-length dress with sophisticated details, such as asymmetrical hemline, tailored cuts with emphasis on the waist and decorative belts.
Likewise, in another example of querying ``sweet'' cardigans and ``edgy'' cardigans: the former retrieves cardigans featuring a looser fit, particularly in the sleeves; while results from the latter are mostly dark-colored with tight sleeves.

